\documentclass[final,5p,times,twocolumn]{elsarticle}

\usepackage{amssymb}

\usepackage{algorithm}
\usepackage{algorithmic}
\usepackage{amsmath}
\usepackage{amsfonts}
\usepackage{booktabs}
\usepackage[table,xcdraw]{xcolor}
\usepackage{multirow}

\newcommand{\mbf}[1]{\mathbf{#1}}
\newcommand{\mbb}[1]{\mathbb{#1}}
\newcommand{\mca}[1]{\mathcal{#1}}

\newcommand{\deriv}[1]{\, \mathrm{d} #1 }
\newcommand\figcaption{\def\@captype{figure}\caption} 
\newcommand\tabcaption{\def\@captype{table}\caption} 
\usepackage[hyphens]{url}

\begin{document}

\begin{frontmatter}



\title{Uncertainty-Aware Pedestrian Trajectory Prediction via Distributional Diffusion}

\author[label1,label2]{Yao Liu\corref{cor}\fnref{equ}}
\ead{y.liu@mq.edu.au}

\author[label2]{Zesheng Ye\fnref{equ}}
\ead{zesheng.ye@unsw.edu.au}

\author[label4]{Rui Wang}
\ead{wangrui@ise.neu.edu.cn}

\author[label5]{Binghao Li}
\ead{binghao.li@unsw.edu.au}

\author[label1]{Quan Z. Sheng}
\ead{michael.sheng@mq.edu.au}

\author[label2,label6]{Lina Yao}
\ead{lina.yao@unsw.edu.au}

\affiliation[label1]{organization={School of Computing, Macquarie University},
            city={Sydney},
            country={Australia}}

\affiliation[label2]{organization={School of Computer Science and Engineering, University of New South Wales},
            city={Sydney},
            country={Australia}}


\affiliation[label4]{organization={College of Information Science and Engineering, Northeastern University},
            city={Shenyang},
            country={China}}
            
\affiliation[label5]{organization={School of Minerals and Energy Resources Engineering, University of New South Wales},
            city={Sydney},
            country={Australia}}
            
\affiliation[label6]{organization={Data 61, CSIRO},
            city={Sydney},
            country={Australia}}

\cortext[cor]{Corresponding author.}
\fntext[equ]{Both authors contributed equally to this research.}

\begin{abstract}
Tremendous efforts have been put forth on predicting pedestrian trajectory with generative models to accommodate uncertainty and multi-modality in human behaviors.
An individual's inherent uncertainty, e.g., change of destination, can be masked by complex patterns resulting from the movements of interacting pedestrians.
However, latent variable-based generative models often entangle such uncertainty with complexity, leading to limited either latent expressivity or predictive diversity.
In this work, we propose to separately model these two factors by implicitly deriving a flexible latent representation to capture intricate pedestrian movements, while integrating predictive uncertainty of individuals with explicit bivariate Gaussian mixture densities over their future locations.
More specifically, we present a model-agnostic uncertainty-aware pedestrian trajectory prediction framework, parameterizing sufficient statistics for the mixture of Gaussians that jointly comprise the multi-modal trajectories.
We further estimate these parameters of interest by approximating a denoising process that progressively recovers pedestrian movements from noise.
Unlike previous studies, we translate the predictive stochasticity to explicit distributions, allowing it to readily generate plausible future trajectories indicating individuals' self-uncertainty.
Moreover, our framework is compatible with different neural net architectures.
We empirically show the performance gains over state-of-the-art even with lighter backbones, across most scenes on two public benchmarks.

\end{abstract}










\end{frontmatter}



\section{Introduction}

Predicting human movements and pedestrian trajectories forms the cornerstone of safe autonomous human-machine interaction systems, such as intelligent transportation and underground mine automation~\cite{kbs_traffic1, kbs_traffic2, mine_bg3}.
In areas with congested intersections, significant safety risks can emerge; this shall be mitigated by detecting people's movements.
However, this task is challenged by highly uncertain human behaviors resulting from interleaved self- and inter-pedestrian uncertainties.
From one perspective, humans exhibit a subjectivity that cannot be fully captured through historical trajectories. 
Even when sharing identical historical trajectories, individuals may not necessarily follow the same future trajectories. 
Additionally, since pedestrians are unable to determine each other's intentions and destinations, they must maintain a level of spatial distance. 
This requirement adds to the overall unpredictability of the system.
These uncertainties highlight the complex and multi-modal nature of human motion~\cite{gupta2018social, sun2021three}. 
This complex and multi-modal nature is caused by the self-uncertainty of pedestrians; hence, in the field of trajectory prediction, it's impossible to estimate with zero-error accuracy. 
This characteristic, where one historical trajectory corresponds to multiple potential future trajectories, is termed "multi-modal". 
Pedestrians may have multiple plausible futures due to inherent uncertainties. 
Current research typically predicts multiple plausible future trajectories for pedestrians, with the expectation that the true future trajectory will align with one of them.

In terms of complexity, trajectory prediction is viewed as a sequence-to-sequence task from a spatial-temporal viewpoint.
In particular, the spatial correlation among pedestrians is attributed to the result of social factors~\cite{helbing1995social, alahi2016social, liu2021social}.
For example, people also pay attention to avoid collisions while walking closely with companions.
This motivates the adoption of increasingly potent non-linear function approximators to represent complex spatial-temporal correlations when further taking into account temporal dependencies, most notably importance-weighted neighboring aggregation~\cite{liang2019peeking, sadeghian2019sophie} with Transformers~\cite{vaswani2017attention, kbs_trans, KBS_bert} and dynamic social-temporal graphs~\cite{huang2019stgat, salzmann2020trajectron++, KBS_graph}.

Upon this, it is now a standard practice to forecast multi-modal trajectories using generative models with latent variables~\cite{gupta2018social}, whereby the plausibility of future trajectories can hopefully be covered by a predictive distribution conditioned on historical context.
By harnessing latent stochasticity, models estimate the distribution~(either explicitly or implicitly) of possible trajectories to approach the uncertainty.
Often, methods with explicit predictive density functions mostly rely on conditional variational autoencoders~(VAEs)~\cite{ivanovic2019trajectron, mohamed2020social, chai2019multipath, chen2021personalized} that assume bivariate Gaussian outputs.
Yet, since multi-agent trajectory data often span complex, high-dimensional spaces, VAEs could suffer from posterior collapse as a result of mismatched data and simple prior distributions~\cite{tomczak2018vae}, eventually leading to limited latent expressivity and unnatural generations~\cite{gu2022stochastic}.
In contrast, those implicit counterparts~\cite{gupta2018social, sadeghian2019sophie} could benefit from more flexible generations without setting a fixed density form beforehand, thanks to the adoption of Generative Adversarial Networks~(GANs)~\cite{goodfellow2020generative}.
Still, training GANs has proven difficult to mitigate mode collapse~\cite{gulrajani2017improved, radford2015unsupervised} and unstable gradients~\cite{karnewar2020msg, kodali2017convergence}.
These latent variable models may struggle to balance model expressivity with multi-modality, given the entanglement between data complexity and task-inherent uncertainty.

\begin{figure}
    \begin{center}
        \begin{small}
            \includegraphics[width=0.9\linewidth]{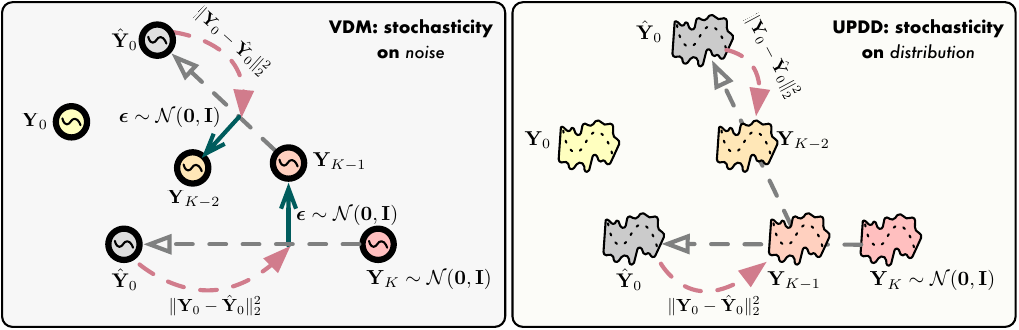}
        \end{small}
    \end{center}
    \caption{Illustrative between VDMs\protect\footnotemark[1] and UPDD. While VDMs rely on random noises throughout diffusion to introduce stochasticity, UPDD inherently includes predictive uncertainty by modeling distributions over trajectories. }
    \label{fig: motivation}
\end{figure}

\footnotetext[1]{We term the direct use of diffusion models to generate trajectory points as vanilla trajectory diffusion models (VDMs).}

A recent bloom has been witnessed in denoising diffusion probabilistic models~(i.e., diffusion models)~\cite{NEURIPS2020_4c5bcfec} applied to image and text generation~\cite{rombach2022high, austin2021structured}.
Diffusion models draw inspiration from non-equilibrium thermodynamics that simulates particle diffusion by progressively injecting noise into data until the signal is destroyed.
Then sampling high-fidelity data from random noise is achievable by learning to reverse diffusion~\cite{sohl2015deep}.
This setting lends itself well to modeling pedestrian motion intuitively:
As pedestrian movements within a system are determined by certain constraints, e.g., destinations and social regulations, adding noise can be viewed as discarding these constraints, resulting in more indeterminate human behaviors.
In due course, the trajectories will become completely unpredictable once the system behaves like random noise.
One can recover the true trajectory density from a noisy system by approximating the reverse of the aforementioned diffusion process.
In light of this, diffusion models were previously applied to trajectory prediction as attempted in trajectory diffusion models.
For instance, MID~\cite{gu2022stochastic} assumes the indeterminacy of pedestrians' movements can be simulated by a Markovian diffusion process resembling our motivation.
However, given that diffusion models limit the latent dimension to the input data, VDMs must output the exact coordinates of every trajectory with a complete single reverse process.
When multi-modal prediction is involved, this reverse process must be repeated multiple times with different sampling noise, not to mention the multiple iterations required to generate even a single sample.
In this sense, MID is inefficient for being much slower than previous studies~(roughly $\times 39$ slower than~\cite{salzmann2020trajectron++}).

Therefore, we propose explicitly parameterizing the predictive distribution of trajectories, rather than the coordinates of the trajectories themselves, as illustratedin Fig.~\ref{fig: motivation}.
Consequently, an individual's self-uncertainty can be separated from modeling the complex interactions of multiple pedestrians. 
Our model is motivated by the computation of sufficient statistics for the trajectory distribution, achieved through the assumption of a bivariate Gaussian distribution, which is then applied to diffusion models. 
Subsequently, we are able to predict the future distribution of trajectories across all pedestrians at future time steps based on the joint parametric distribution.
By doing this, the reliance on randomness shifts from noise to the predictive distribution, with each reverse diffusion process obtaining the distribution of trajectories. This enables us to easily sample multiple trajectories from the distribution, thereby avoiding the computational burden of performing reverse diffusion for each trajectory.

In more detail, our first step maps trajectory data into sufficient statistics for a predictive trajectory distribution.
Following, we encode historical and neighboring information of each pedestrian, offering guidance information~\cite{ho2022classifier} to a conditional diffusion model-based trajectory generator\footnote[2]{There might be ambiguity given by the naming of {\it trajectory generator}. More precisely, we generate sufficient statistics, i.e., the distribution of trajectory, then we sample trajectories therefrom.}, an approximation of the true reverse diffusion.
In addition, concerning the efficiency issue of VDMs, we follow~\cite{song2020denoising} that relaxes the Markovian assumption of DDPM~\cite{gu2022stochastic} for faster sampling noise sampling, through adding multiple Markov jumps at once with deterministic transitions.
Since our model generates a trajectory distribution, the loss of stochasticity in the DDIM scheme is negligible.
The primary distinction between our model and VDMs lies in the diffusion of the sufficient statistics of the trajectory distribution, rather than the trajectories directly. 
This approach offers the advantage of segregating the self-uncertainty of pedestrians, facilitating efficient trajectory sampling through the trajectory distribution.

In summary, our contributions are:
\begin{itemize}
\item We present an {\bf U}ncertainty-aware trajectory {\bf P}rediction framework through {\bf D}istributional {\bf D}iffusion~(UPDD). UPDD parameterizes the multi-modal predictive trajectory distribution by approximating reverse diffusion transitions, which progressively improves the predictive confidence of future locations.
\item We separate uncertainty estimation of individual behaviors from complex multi-agent motions by using diffusion models applied with explicit bivariate Gaussian density.
UPDD generalizes prior models by incorporating predictive uncertainty, enabling multi-modal sampling from trajectory distributions with just one reverse process. Furthermore, independently modeling pedestrians' intrinsic uncertainty facilitates the acceleration of the iterative denoising process through the DDIM paradigm applied to Diffusion.
\item We show that UPDD empirically outperforms state-of-the-art on different trajectory prediction benchmarks, pronounced with superior efficiency over VDM approaches.
\end{itemize}

\section{Related Work}
\subsection{Trajectory Forecasting as Spatial-temporal Prediction}
Often referred to as a sequence-to-sequence task, pedestrian trajectory forecasting uses historical trajectories as input to predict the multi-agent locations along future timesteps, taking their spatial correlations into account.
Social-LSTM~\cite{alahi2016social} is a pioneering work that incorporates social interactions between agents into a recurrent neural network~(RNN) to capture both spatial and temporal dependencies.
Later studies focus mainly on encoding dynamic influences from neighbors effectively.
The attention-based methods~\cite{vemula2018social, liang2019peeking, gupta2018social, sadeghian2019sophie} assign different importance to aggregated neighbors, while others construct a spatial-temporal graph to represent social interactions~\cite{zhang2019sr, huang2019stgat, ivanovic2019trajectron, salzmann2020trajectron++}.
Nonetheless, RNNs are often faulted for failing on parallelism.
This motivates the use of Transformer~\cite{vaswani2017attention} as an alternative.
For instance, STAR~\cite{yu2020spatio} employs both temporal and spatial Transformers to capture complex spatial-temporal interactions in each dimension, respectively.
AgentFormer~\cite{yuan2021agentformer} proposes modeling two dimensions concurrently using a Transformer to prevent potential information loss caused by the independent encoding of either dimension.
Even so, Transformers can sometimes pose challenges to model efficiency due to their parameter-intensive characteristics~\cite{shi2021sgcn}.
Instead, we discuss an alternative with both spatial and temporal convolutions toward a parameter-efficient solution of trajectory prediction.
We empirically demonstrate that one can choose to use lightweight backbone networks, provided that a flexible generative framework is effectively employed.

\subsection{Trajectory Forecasting as Generative Modeling}
To account for the uncertainty and multi-modality of future dynamics, recent studies have embraced generative models to predict pedestrian trajectory, conditioned on intermediate representations of historical trajectories.
One can be categorized into either type depending on whether the model outputs an explicit predictive distribution.
Centering around conditional VAEs,~\cite{lee2017desire, ivanovic2019trajectron, mohamed2020social, chai2019multipath, chen2021personalized} learn to maximize the likelihood of predictive distribution with explicit function forms.
As an example,~\cite{mohamed2020social, chai2019multipath} assume the locations subject to a bi-variate Gaussian at each timestep.
While explicit density allows for easier sampling of multi-modal generation, it comes at the cost of limited latent expressivity and jagged trajectory generation~\cite{gu2022stochastic} due to posterior collapse during optimization~\cite{tomczak2018vae}.

The generation with implicit density~\cite{gupta2018social, sadeghian2019sophie, zhao2019multi, dendorfer2021mg}, on the other hand, is shown to be more flexible by following GANs~\cite{goodfellow2020generative}.
These methods generate trajectories from random noise and adopt a discriminator to distinguish generated ones from the ground truth.
However, training GAN is notorious for being unstable, despite efforts to alleviate mode collapses, such as reversible transformations~\cite{huang2019stgat} and multiple generators~\cite{dendorfer2021mg}.
Later on, trajectory diffusion models~(VDMs) shed light on addressing issues in both VAEs and GANs worlds.
The first attempt MID~\cite{gu2022stochastic} applies diffusion models~\cite{NEURIPS2020_4c5bcfec} to predict future coordinates.
Similar to our motivation, MID gradually reduces the predictive indeterminacy by learning to reverse a diffusion process; 
this provides MID with more flexibility than VAE-based explicit density models, while being easier to be trained than GAN-based approaches.
Nevertheless, MID outputs exact coordinates, essentially indicating that predictive stochasticity and multi-modality hinge on sampling diverse random Gaussian noise before initiating the reverse diffusion process.
In this case, obtaining multiple plausible paths is time-consuming since each sample can only be obtained with a complete denoising process.
Unlike MID, our method enforces a mixture of bi-variate Gaussian predictions per timestep, thus enabling faster sampling from the explicit density function while still handling complex patterns of multi-pedestrian movements.


\section{Methodology}

\subsection{Problem Formulation}
Assuming $N$ co-existing pedestrians walking in the scene, we seek to predict future pedestrian trajectory without auxiliary information, e.g., scene context and destination.
Having observed their locations $\mathbf{X} = \{ \mathbf{x}^{(n)} \}_{n=1:N} \in \mathbb{R}^{N \times T \times 2}$ with $\mathbf{x}^{(n)} = \{ \boldsymbol{x}^{(n)}_{t} \}_{t=1:T}$ throughout $T$ history timesteps, we generate multiple plausible trajectories for each pedestrian indicating multi-modal future locations.
Concerning the entire scene, we predict $\hat{\mathbf{Y}} = \{ \mathbf{y}^{(n)} \}_{n=1:N} \in \mathbb{R}^{N \times T^{\prime} \times 2} $ in the next $T^{\prime}$ timesteps by parameterizing a conditional predictive distribution $p_{\theta}(\mathbf{Y} | \mathbf{X})$.
Since $p_{\theta}(\mathbf{Y} | \mathbf{X})$ describes a joint outcome over $N \times T^{\prime}$ variables, we can sample multiple predicted trajectories for any pedestrian therefrom.

\subsection{Method Overview}
As part of the generative modeling, we aim to recover the true distribution of future trajectories
$p(\mathbf{Y} | \mathbf{X})$ with a parametric approximation $p_{\theta}(\mathbf{Y} | \mathbf{X})$.
For each pedestrian~(indexed by~$n$), let $p(\boldsymbol{y}_{n}^{t} | \boldsymbol{y}_{n}^{t-1}, \mathbf{X}) = \mathcal{N}(\boldsymbol{y}_{n}^{t}; \mu_1, \mu_2, \sigma_1, \sigma_2, \rho)$ be a bi-variate Gaussian parameterized by its sufficient statistics, where $\mu_1, \mu_2, \sigma_1, \sigma_2$ are predictive means and standard deviations for each 2D coordinate, $\rho$ measures the correlation between two dimensions.
To capture multi-modal $p(\mathbf{Y} | \mathbf{X})$, we build our generative process upon diffusion models, equipped with explicit density forms at each location\footnote[3]{We only assume the bi-variate Gaussian density in each location, but not the concrete form of $p_{\theta}(\mathbf{Y} | \mathbf{X})$.}.
Specifically, we estimate predictive statistics by approximating a reverse diffusion chain that starts from random noise and progressively reduces the uncertainty of possible future trajectories.

\begin{figure*}
    \begin{center}
        \begin{small}
            \includegraphics[width=0.9\textwidth]{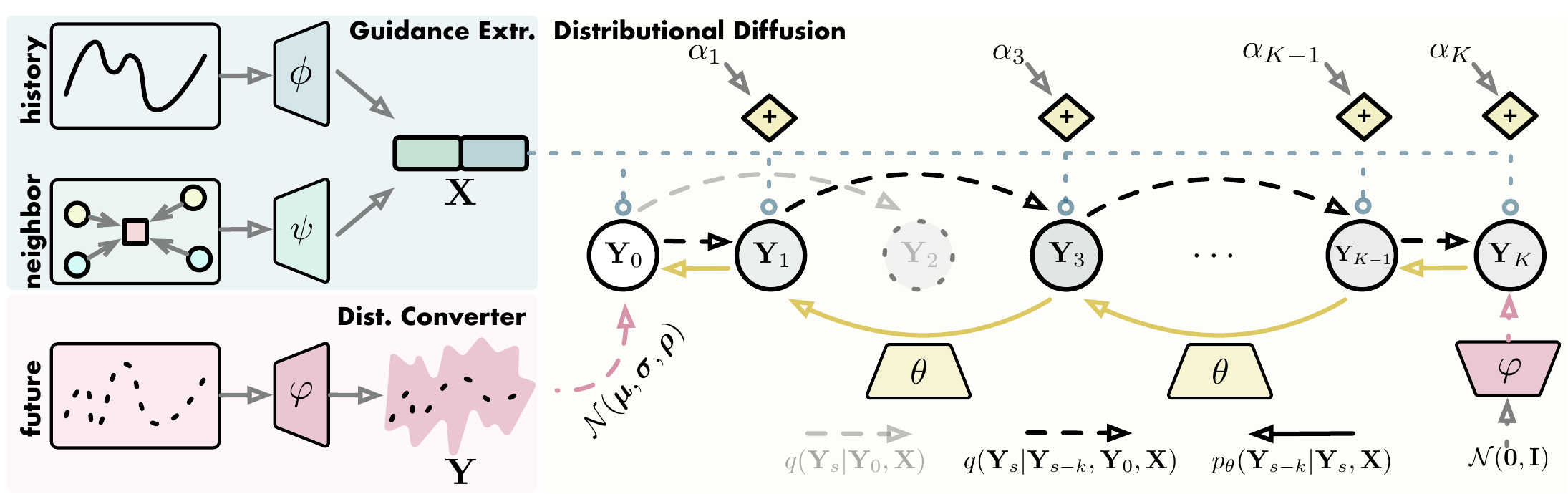}
        \end{small}
    \end{center}
    \caption{
    The overview of UPDD, composed of {\color[HTML]{A1CDB3} {\it guidance extractor}}, {\color[HTML]{FFB6C1} {\it distribution converter}} and {\color[HTML]{FFBF00} {\it distributional diffusion}}.
    To account for multi-modal human movements, UPDD approximates the distributions of future trajectories under a diffusion model-based generative framework.
    We encode historic and neighboring effects for each pedestrian to guide conditional diffusion;
    we map future coordinates into sufficient statistics of a parametric density estimation, enabling fast sampling of future trajectories therefrom;
    we also speed up the generation with a non-Markovian ``diffusion'' chain by skipping certain steps.
    } 
    \label{fig: overall}
\end{figure*}

Fig.~\ref{fig: overall} overviews the proposed Uncertainty-Aware trajectory Prediction framework with Distributional Diffusion~(UPDD).
UPDD includes four trainable modules $\Theta := \{ \phi, \psi, \varphi, \theta \}$ and three functional components.
(a) {\it guidance information extractor} is applied to each pedestrian, encoding historical trajectory with $\phi(\cdot)$ and social information with $\psi(\cdot)$, respectively;
(b) {\it distribution converter} $\varphi(\cdot)$ establishes a mapping from exact future trajectories to the Gaussian parameters;
(c) {\it trajectory generator} $\theta(\cdot)$ approximates the true denoising transition and parameterizes the distribution of future trajectory.
These trainable modules are jointly optimized under a supervised paradigm.

Our model uses a framework based on the diffusion process, where the observed trajectories serve as the guiding information to lead the diffusion and reverse diffusion process. 
In the forward process we diffuse the sufficient statistics of the future trajectory until Gaussian noise, and in the reverse process we discard the indeterminacy of the sufficient statistics of the sample from Gaussian noise until we recover the sufficient statistics of the trajectory.
Guidance information includes historical information and social interaction information.
The historical information is extracted from the observed trajectories and it represents the temporal features of the current pedestrian itself.
The social interaction information is aggregated from the historical information of the neighbors around the current pedestrian, which represents the spatial features of the current pedestrian.
Pedestrians at the same time are neighbors of each other, and the number of pedestrians at each time point may not be consistent.
In the diffusion process, we deal with the sufficient statistics of the trajectory rather than the trajectory itself, thus preserving the subjective uncertainty of the pedestrian; in the reverse diffusion process, we discard the indeterminacy of the sufficient statistics of the noise sample, which is the stochastic nature of the motion and it is the content that the model needs to learn, and finally generate the sufficient statistics of the trajectory, which contains the subjective uncertainty of the pedestrian that should be preserved.

\subsection{Extracting Guidance Information}
\label{sec: guidance}
The observations of raw trajectory describe the relative coordinates of pedestrians.
To enrich input features, we also compute velocity and acceleration solely based on coordinates.
Commonsense says that the trajectories of pedestrians depend on their past movements and surroundings.
In UPDD, we explore two pure CNN-based encoders $\phi(\cdot)$ and $\psi(\cdot)$ for extracting each pedestrian's historical and neighboring information.
Nonetheless, one can substitute them with any RNN-based, e.g.,~\cite{gupta2018social} or Transformer-based, e.g.,~\cite{yuan2021agentformer}, spatial-temporal encoders, since UPDD is essentially a model-agnostic framework.

\subsubsection{Encoding Historical Information.}
The raw trajectory data describes the relative coordinates of pedestrians.
Concretely, the history encoder $\phi(\cdot)$ performs the following operations.
For each pedestrian $n$ with historical observations of $T$ steps, we first apply a 1D-CNN with $T$ kernels on the input features to calculate the influences of previous steps on the last~(current) timestep.
Since the step at $t=T$ holds the most influence over $t=T+1$, we concatenate the last timesteps from the resulting $T$ feature maps, and weigh the concatenation using self-attention~\cite{vaswani2017attention} to obtain the historical context encoding $\hat{\mathbf{x}}^{(n)}_{\phi}$.

\subsubsection{Encoding Neighboring Influence.}
The neighboring effects are typically considered substantial for each pedestrian, referred to as social interaction information~\cite{alahi2016social, sadeghian2019sophie}.
Thus, we design a neighbor encoder $\psi(\cdot)$ to encode the neighbors' history information and its effects on each pedestrian.
As the number of neighbors may vary over time, we first aggregate the neighbors' features for each timestep $t$, leading to $T$ aggregated input features of pedestrian $n$.
Accordingly, the subsequent operations resemble $\phi(\cdot)$ but handle the neighbors' features, eventually producing the neighboring context encoding $\hat{\mathbf{x}}^{(n)}_{\psi}$.
We include more details of $\phi(\cdot)$ and $\psi(\cdot)$ in Sec.~\ref{sec: arch_details}.

Having obtained both historical and neighboring information $\hat{\mathbf{x}}^{(n)}_{\phi}, \hat{\mathbf{x}}^{(n)}_{\psi}$, we concatenate them into pedestrian guidance information used in the {\it distributional diffusion}.
In the following sections, we overload the term $\mathbf{X}$ to denote the guidance information of all pedestrians and omit the subscript $\phi, \psi$ for clarity unless mentioned.

\subsection{Mapping Trajectory to Sufficient Statistics}
\label{sec: mapping}
To preserve the predictive uncertainty of future trajectories, we map the exact coordinates into sufficient statistics of the bi-variate Gaussian density function at each timestep, i.e., $\varphi: \mathbb{R}^{N \times T^{\prime} \times 2} \to \mathbb{R}^{N \times T^{\prime} \times 5}$.
In other words, we aim to establish a deterministic non-linear transformation over all samples in the location space.
Our implementation for {\it distribution converter} here is a CNN~(still, can be any function approximators), such that
\begin{equation}
    \begin{aligned}
        \varphi: \boldsymbol{y}_{t}^{(n)} \mapsto \boldsymbol{y}_{t, \varphi}^{(n)} := \left\{ \mu_1, \mu_2, \sigma_1, \sigma_2, \rho \right\} \in \mathbb{R}^{5},
    \forall n, \forall t.
    \end{aligned}
\end{equation}
Each of these 5D vectors determines each pedestrian's coordinate distribution and constitutes the joint event $\mathbf{Y}_{\varphi}$ taking all pedestrians and timesteps into account.
We consider them as ``samples'' in the following diffusion model-based generation process, described in Sec.~\ref{sec: diffusion}.
We discuss the optimization of $\varphi$ in Sec.~\ref{sec: optimization}.

\subsection{Parameterizing Predictive Distribution}
\label{sec: diffusion}
\subsubsection{Diffusion Process}
We now detail the diffusion process for reconstructing future trajectories over all pedestrians with the aforementioned sufficient statistics by considering them as ``samples'' in the generative model.
We overload and slightly abuse the term $\mathbf{Y}_{0}$ to denote these samples for brevity.
As discussed in~\cite{NEURIPS2020_4c5bcfec}, the diffusion process defines a $K$-iteration Markov chain $\{ \mathbf{Y}_0, \dots, \mathbf{Y}_K | \mathbf{X} \}$ corrupting the samples $\mathbf{Y}_0$ to the noises $\mathbf{Y}_K$ that span over the whole walkable area in our case.
The diffusion chain is fixed to $q(\mathbf{Y}_{1:K} | \mathbf{Y}_{0}, \mathbf{X}) = \prod_{k=1}^{K} q(\mathbf{Y}_k | \mathbf{Y}_{k-1}, \mathbf{X})$ in an autoregressive manner, with each intermediate transition defined as a Gaussian, parameterized by strictly decreased $\alpha_{k-1} > \alpha_{k} \in (0, 1]$:
\begin{equation}
    \resizebox{0.6\linewidth}{!}{$
    \begin{aligned}
        q & (\mathbf{Y}_k | \mathbf{Y}_{k-1}, \mathbf{X}) 
        & = \mathcal{N} \left( \mathbf{Y}_k | \mathbf{X}; \sqrt{\frac{\alpha_{k}}{\alpha_{k-1}}} \mathbf{Y}_{k-1}, \left( 1 - \frac{\alpha_{k}}{\alpha_{k-1}} \right) \mathbf{I}  \right).    
    \end{aligned}
    $}
\end{equation}
We can compute a closed-form final state $\mathbf{Y}_K$ using $\mathbf{Y}_0$, enabled by recursive sampling with the reparameterization trick~\cite{kingma2013auto} of Gaussian transitions,
\begin{equation}\label{eq: ddqpm_forward_final}
    \resizebox{0.5\linewidth}{!}{$
    \begin{aligned}
        q(\mathbf{Y}_K | \mathbf{Y}_{0}, \mathbf{X}) & = \int q( \mathbf{Y}_{1:K} | \mathbf{Y}_{0}, \mathbf{X}) \, \mathrm{d} \mathbf{Y}_{1:K-1}  \\
                                   & = \mathcal{N} \left( \mathbf{Y}_K | \mathbf{X}; \sqrt{\alpha_K} \mathbf{Y}_0, \left(1 - \alpha_K \right) \mathbf{I} \right).
    \end{aligned}
    $}
\end{equation}
That is, by sequentially adding noise to $\mathbf{Y}_{0}$, the forward diffusion eventually converts it into a standard Gaussian noise $\mathbf{Y}_{K}$ when $\alpha_K$ is sufficiently close to 0.
Learning to reverse the forward process allows us to recover input samples from noise.
With the guidance information extracted from historical observations $\mathbf{X}$, we thus have the conditional generation process by marginalizing all the intermediate variables,
\begin{equation}
    \resizebox{0.5\linewidth}{!}{$
    \begin{aligned}
        p_{\theta}(\mathbf{Y}_0 | \mathbf{X}) & = \int p_{\theta} (\mathbf{Y}_{0:K} | \mathbf{X}) \, \mathrm{d} \mathbf{Y}_{1:K} \\
        & = \int p_{\theta}(\mathbf{Y}_{K})  \prod_{k=1}^{K} p_{\theta}(\mathbf{Y}_{k-1} | \mathbf{Y}_{k}, \mathbf{X} ) \, \mathrm{d} \mathbf{Y}_{1:K} 
    \end{aligned}
    $}
\end{equation}
with $p_{\theta}(\mathbf{Y}_{K}) = \mathcal{N}(0, \mathbf{I})$.
The parameters of $\theta$ are shared throughout the reverse chain, implemented as {\it trajectory generator}.
Since the true reverse process $q(\mathbf{Y}_{k-1} | \mathbf{Y}_{k}, \mathbf{X})$ is intractable, we optimize a variational lower bound instead:
\begin{equation}\label{eq: ddpm_elbo}
    \resizebox{0.8\linewidth}{!}{$
    \begin{aligned}
        & \max_{\theta} \,\mathbb{E}_{p(\mathbf{Y}_0 | \mathbf{X})} \left[ \log p_{\theta}(\mathbf{Y}_0 | \mathbf{X}) \right]  \\
        & \geq \max_{\theta} \mathbb{E}_{p(\mathbf{Y}_{0}, \mathbf{Y}_{1}, \dots, \mathbf{Y}_{K} | \mathbf{X})} \left[ \log \frac{ p_{\theta} (\mathbf{Y}_{0:K} | \mathbf{X}) }{ q(\mathbf{Y}_{1:K} | \mathbf{Y}_{0}, \mathbf{X}) } \right] \\
        & = \max_{\theta} \mathbb{E}_{p(\mathbf{Y}_{0}, \mathbf{Y}_{1}, \dots, \mathbf{Y}_{K}  | \mathbf{X})}  
         \quad \left[ \log p_{\theta} (\mathbf{Y}_{K}) - \sum_{k=1}^{K} \log \frac{ p_{\theta} ( \mathbf{Y}_{k-1} | \mathbf{Y}_{k}, \mathbf{X} ) }{ q(\mathbf{Y}_{K} | \mathbf{Y}_{k-1}, \mathbf{X}) }  \right],
    \end{aligned}
    $}
\end{equation}
where the true distribution is termed by $p(\mathbf{Y}_0 | \mathbf{X})$ to be fitted with $p_{\theta}(\mathbf{Y}_0 | \mathbf{X})$.
Empirically, Eq.~\ref{eq: ddpm_elbo} can be simplified to the mean squared error between the ``predicted'' noise added to the sample and a noise $\epsilon_k$ randomly drawn from a standard diagonal Gaussian throughout $K$ steps,
\begin{equation}\label{eq: ddpm_elbo_emp}
    \begin{aligned}
        \min_{\Theta} \mathcal{L}_{\text{emp}} & = \sum_{k=1}^{K} \mathbb{E}_{\mathbf{Y}_0 \sim p(\mathbf{Y}_0 | \mathbf{X}), \epsilon_k \sim \mathcal{N}(0, \mathbf{I})} \left[ || \theta( \hat{\epsilon}_k ) - \epsilon_k ||_2^2 \right], \\
    \end{aligned}
\end{equation}
with $\hat{\epsilon}_k = \sqrt{\alpha_k} \mathbf{Y}_0 + \sqrt{1 - \alpha_k} \epsilon_k$.
$\mathbf{Y}_k$ is sampled by applying the reparameterization trick on $q(\mathbf{Y}_k | \mathbf{Y}_0, \mathbf{X})$ each step\footnote[4]{While Eq.~\ref{eq: ddpm_elbo} and Eq.~\ref{eq: ddpm_elbo_emp} are designed for $\theta$, the variables $\mathbf{Y}_0$ and $\mathbf{X}$ are parameterized by $\varphi$ and $\phi, \psi$, so they are jointly optimized.}.

Moreover, we could adopt the approach suggested in~\cite{song2020denoising} that ``shorten'' the Markovian diffusion process by adding certain Markov jumps simultaneously, as a way to expedite sampling from diffusion models with a similar surrogate objective.

\subsubsection{Non-Markovian Forward Process}
\label{sec: non_markov_fp}
Observing a large~$K$ in Eq.~\ref{eq: ddqpm_forward_final} introduces a trade-off between how close~$p(\mathbf{Y}_K)$ gets to a standard Gaussian~\cite{sohl2015deep} and the efficiency of autoregressive sampling, we employ a non-Markovian forward process that generalizes a Markovian diffusion while maintaining Gaussianity in Eq.~\ref{eq: ddqpm_forward_final}.
Consider a forward process where the transition also depends on $\mathbf{Y}_0$, we have its joint distribution as,
\begin{equation}
    \resizebox{0.6\linewidth}{!}{$
    \begin{aligned}
        q_{\gamma}  (\mathbf{Y}_{1:K} | \mathbf{Y}_0, \mathbf{X}) 
                    = q_{\gamma} ( \mathbf{Y}_{K} | \mathbf{Y}_{0}, \mathbf{X}) \prod_{k = 2}^{K} q_{\gamma} ( \mathbf{Y}_{k-1} | \mathbf{Y}_{k}, \mathbf{Y}_{0}, \mathbf{X}).
    \end{aligned}
    $}
\end{equation}
To ensure Gaussianity in each intermediate state depending on $\mathbf{Y}_0$, such that
\begin{equation}
\resizebox{0.5\linewidth}{!}{$
q_{\gamma}( \mathbf{Y}_{k} | \mathbf{Y}_{0}, \mathbf{X}) =\mathcal{N}(\mathbf{Y}_{k} | \mathbf{X}; \sqrt{\alpha_k} \mathbf{Y}_0, (1 - \alpha_k) \mathbf{I})
$}
\end{equation}
for all $k > 0$, we define the true reverse transition as
\begin{equation}\label{eq: non_mk_denoise}
    \resizebox{0.8\linewidth}{!}{$
    \begin{aligned}
         q_{\gamma} ( \mathbf{Y}_{k-1} | \mathbf{Y}_{k}, \mathbf{Y}_{0}, \mathbf{X}) = 
         \mathcal{N}\left( \sqrt{\alpha_{k-1}} \mathbf{Y}_0 + \sqrt{1 - \alpha_{k-1} - \gamma_k^2} \frac{ \mathbf{Y}_k - \sqrt{\alpha_k} \mathbf{Y}_0 }{ \sqrt{ 1 - \alpha_k} }, \gamma_k^{2} \mathbf{I} \right),
    \end{aligned}
    $}
\end{equation}
where $\gamma \in \mathbb{R}^{K}$ controls the stochasticity added to each step.
That is, the transition becomes deterministic if $\gamma \to 0$.
Resembling Eq.~\ref{eq: ddpm_elbo}, the variational lower bound is empirically approximated by,
\begin{equation}~\label{eq: ddim_elbo}
    \resizebox{0.6\linewidth}{!}{$
    \begin{aligned}
        & \mathbb{E}_{q_\gamma(\mathbf{Y}_{0:T})} \left[  \log \frac{p_{\theta}(\mathbf{Y}_{0:K} | \mathbf{X})}{ q_{\gamma}(\mathbf{Y}_{1:K} | \mathbf{Y}_{0}, \mathbf{X}) } \right] \\
         = & \mathbb{E}_{q_\gamma(\mathbf{Y}_{0:T})} \left[  \log p_{\theta} (\mathbf{Y}_K) + \sum_{k=1}^{K} \log p_{\theta} ( \mathbf{Y}_{k-1} | \mathbf{Y}_{k}, \mathbf{X} )  \right. \\ 
        & \quad \left. - \log q_{\gamma} (\mathbf{Y}_{K} | \mathbf{Y}_{0}, \mathbf{X}) - \sum_{k=2}^{K} (\log q_{\gamma} ( \mathbf{Y}_{k-1} | \mathbf{Y}_{k}, \mathbf{Y}_{0}, \mathbf{X} ) \right],
    \end{aligned}
    $}
\end{equation}
where $\log p_{\theta} ( \mathbf{Y}_{k-1} | \mathbf{Y}_{k}, \mathbf{X} )$ approximates the true denoising step and is parameterized by our {\it trajectory generator}, whereas the other three terms have been derived previously.
Still, the gradients are calculated by mean square error as in Eq.~\ref{eq: ddpm_elbo_emp}.
More precisely, the generation starts from noise $p_{\theta}(\mathbf{Y}_K) = \mathcal{N}(0, \mathbf{I})$ and sequentially parameterizes next transition by replacing $\mathbf{Y}_0$ in Eq.~\ref{eq: non_mk_denoise}:
\begin{equation}
    \resizebox{0.7\linewidth}{!}{$
    \begin{aligned}
        p_{\theta}(\mathbf{Y}_{k-1}  | \mathbf{Y}_{k}, \mathbf{X})  
         = q_{\gamma} \left( \mathbf{Y}_{k-1} | \mathbf{Y}_{k}, \frac{\mathbf{Y}_k - \sqrt{1 - \alpha_k} \cdot \theta(\mathbf{Y}_{k} | \mathbf{X})}{\sqrt{\alpha_k}} \right),
    \end{aligned}
    $}
\end{equation}
where $\theta(\mathbf{Y}_{k} | \mathbf{X})$ is our {\it trajectory generator} conditioned on guide information $\mathbf{X}$.

\subsubsection{Accelerated Deterministic Reverse Process}
As we have shown, a non-Markovian forward process that conditions each intermediate step on $\mathbf{Y}_0$ allows us to express the joint $q_{\gamma}(\mathbf{Y}_{0:T} | \mathbf{Y}_0)$ in a flexible way, so long as the Gaussianity in $q_{\gamma}(\mathbf{Y}_{t} | \mathbf{Y}_0)$ for all $t > 0$ is ensured.
This is a key takeaway concerning reducing the number of iterations $K$ in the reverse process.
Intuitively, this implies reducing the number of iterations with another forward process fulfilling such requirements.
Specifically, we can arbitrarily sample a sub-sequence $\tau \subset \{1, \dots, K \}$ with length $S < T$ to form a forward process with $q_{\gamma}(\mathbf{Y}_{\tau_{i}} | \mathbf{Y}_0) = \mathcal{N}(\mathbf{Y}_{\tau_{i}}; \sqrt{\alpha_{\tau_i}} \mathbf{Y}_0, \sqrt{1 - \alpha_{\tau_i}} \mathbf{I})$.
As such, the joint distribution can be factorized into
\begin{equation}\label{eq: ddim_joint}
    \resizebox{0.7\linewidth}{!}{$
    \begin{aligned}
         q_{\gamma}(\mathbf{Y}_{1:K} | \mathbf{Y}_0) 
         q_{\gamma}(\mathbf{Y}_{\tau_S} | \mathbf{Y}_0) \prod_{i \in \tau_{S}} q_{\gamma}(\mathbf{Y}_{\tau_{i-1}} | \mathbf{Y}_{\tau_{i}}, \mathbf{Y}_0) \prod_{j \in \bar{\tau}} q_{\gamma}( \mathbf{Y}_j | \mathbf{Y}_0),
    \end{aligned}
    $}
\end{equation}
where $\bar{\tau}$ involves steps not included in $\tau$, thereby decreasing the steps of forward iterations from $K$ to $S$.
Reversely, the generative process is also shortened to $S$ steps.
Having derived each term's closed form in Sec.~\ref{sec: non_markov_fp}, we optimize the reverse process defined by Eq.~\ref{eq: ddim_joint} with Eq.~\ref{eq: ddim_elbo}.
Further, by sampling noise $\mathbf{Y}_{S} \sim p_{\theta}(\mathbf{Y}_{S})$, we get a deterministic generative process by setting $\gamma = 0$, proven to be more semantically meaningful in the latent space~\cite{song2020denoising}.
Notably, setting $\gamma = 0$ means no stochasticity is involved by reverse diffusion - VDMs must rely on the randomness of noise sampling outside the diffusion.

\subsection{Sampling from Predictive Distribution}
From $p(\mathbf{Y}_S)$, one can generate $\hat{\mathbf{Y}}_0$ by running the approximate reverse transition $p_{\theta}(\mathbf{Y}_{k-1} | \mathbf{Y}_{k}, \mathbf{X})$ iteratively.
Recall that we parameterize the explicit density function of all pedestrians at each future timestep, i.e., $\hat{\mathbf{Y}}_0$ being sufficient statistics of the multi-modal predictions.
For any pedestrian $n$, the future trajectory can be generated by sampling from $p_{\theta}(\mathbf{Y}_0 | \mathbf{X})$ throughout every future timestep as
\begin{equation}
\resizebox{0.6\linewidth}{!}{$
    \hat{\mathbf{y}}^{(n)}_{t} := (\hat{\mu}_{t,1}^{(n)}, \hat{\mu}_{t,2}^{(n)}) \sim \mathcal{N} \left(\mu_{t,1}^{(n)}, \mu_{t,2}^{(n)}, \sigma_{t,1}^{(n)}, \sigma_{t,2}^{(n)}, \rho_{t}^{(n)} \right)
$}
\end{equation}
for all $n \in [1, N]$ and $T < t < T^{\prime}$.
In this way, we can further easily and painlessly sample trajectory predictions {\it as much as we want} from even running the reverse chain once.

\subsection{Multi-task Optimization}
\label{sec: optimization}
In addition to Eq.~\ref{eq: ddim_elbo}, we place two objectives to account for the multi-modal nature of trajectories, as well as the consistency of predicted trajectories, respectively.
One, we maximize the log-likelihood of ground-truth future trajectories $\mathbf{Y}$ w.r.t the predictive bi-variate Gaussian parameterized by $\mathbf{Y}_{\varphi}$, towards an improved interpretation of true trajectories,
\begin{equation}\label{eq: loss_llh}
    \resizebox{0.5\linewidth}{!}{$
    \min_{\Theta} \mathcal{L}_{\text{llh}} = - \sum_{n=1}^{N} \sum_{t=T+1}^{T^{\prime}} \log p( \boldsymbol{y}_{t}^{(n)}; \, \boldsymbol{y}_{t, \varphi}^{(n)} ).
    $}
\end{equation}
Second, to ensure that each predictive bi-variate Gaussian matches the correct exact location, we preserve their consistencies by minimizing the mean squared error as
\begin{equation}\label{eq: loss_consistency}
    \resizebox{0.45\linewidth}{!}{$
    \min_{\Theta} \mathcal{L}_{\text{con}} = \sum_{n=1}^{N} || \boldsymbol{y}^{(n)}_{T+1} - \boldsymbol{\mu}^{(n)}_{T+1} ||_2^2.
    $}
\end{equation}
where $\boldsymbol{\mu}^{(n)}_{T+1}$ are predictive means of $\boldsymbol{y}_{\varphi}^{(n)}$ at the first step of predictions, i.e., $t=T+1$.
We then express the overall training objective as a linear combination of Eq.~\ref{eq: ddim_elbo}, Eq.~\ref{eq: loss_llh} and Eq.~\ref{eq: loss_consistency} with trade-off parameters $\lambda_1, \lambda_2$ and $\lambda_3$.

\subsection{Additional Details}

The following Sec.~\ref{sec:add_derivation} includes derivations that, although consistent with established literature, have been modified for terminology consistency with this study. Familiar readers may regard these as supplemental.

\subsubsection{Additional Derivation}
\label{sec:add_derivation}

Most of the derivations below have been discussed in related studies\footnote[5]{especially in~\url{https://calvinyluo.com/2022/08/26/diffusion-tutorial.html} and~\cite{song2020denoising}.}.

\noindent \textbf{Gaussianity of Markovian forward process}.
The forward process of diffusion models can be derived with a Gaussian transition from $q(\mbf{Y}_K | \mbf{Y}_0, \mbf{X})$.
Knowing that each intermediate transition is a Gaussian, we can sample $\mbf{Y}_{k} \sim q(\mbf{Y}_{k} | \mbf{Y}_{k-1}, \mbf{X})$ from a standard Gaussian with the reparameterization trick,
\begin{equation}
\resizebox{0.6\linewidth}{!}{$
    \mbf{Y}_{t} = \sqrt{\beta_{k}} \mbf{Y}_{k-1} + \sqrt{1 - \beta_{k}} \epsilon, \quad \text{with } \epsilon \sim \mca{N}(0, \mbf{I}).
$}
\end{equation}
where 
\begin{equation}
\resizebox{0.4\linewidth}{!}{$
    \alpha_k = \prod_{i=1}^{k} \beta_i \; \Leftrightarrow \; \beta_k = \frac{\alpha_k}{\alpha_{k-1}},
$}
\end{equation}
which has previously\footnote[6]{note the difference between $\alpha_k$ and $\beta_k$. The change in symbol does not change the equivalence. } been defined in Eq.~\ref{eq: ddqpm_forward_final}.
Then the recursive use of the reparameterization trick would eventually lead to the Gaussian of $q(\mbf{Y}_K | \mbf{Y}_0, \mbf{X})$.
To derive $\mbf{Y}_{k}$ for arbitrary $k > 1$, we have
\begin{equation}
\resizebox{0.8\linewidth}{!}{$
    \begin{aligned}
        \mbf{Y}_k & = \sqrt{\beta_k} \mbf{Y}_{k-1} + \sqrt{1-\beta_k} \epsilon^{*}_{k-1}                                                                                                                             \\
                  & = \sqrt{\beta_k} \left( \sqrt{\beta_{k-1}} \mbf{Y}_{k-2} + \sqrt{1-\beta_{k-1}} \epsilon^{*}_{k-2} \right) + \sqrt{1-\beta_k} \epsilon^{*}_{k-1}                                               \\
                  & = \sqrt{\beta_k \beta_{k-1}} \mbf{Y}_{k-2} + \sqrt{\beta_k (1-\beta_{k-1}) + (1-\beta_k) \beta_{k-1}} \epsilon^{*}_{k-2} + \sqrt{1-\beta_k} \epsilon^{*}_{k-1}                              \\
                  & = \underbrace{\sqrt{\beta_k \beta_{k-1}} \mbf{Y}_{k-2} + {\color{blue} \sqrt{1 - \beta_k \beta_{k-1}}} \epsilon_{k-2}}_{\text{recursion}}                                                      \\
                  & = \sqrt{\beta_k \beta_{k-1}} \left( \sqrt{\beta_{k-2}} \mbf{Y}_{k-3} + \sqrt{1 - \beta_{k-2}} \epsilon^{*}_{k-3} \right) + \sqrt{1 - \beta_k \beta_{k-1}} \epsilon_{k-2}                     \\
                  & = \sqrt{\beta_k \beta_{k-1} \beta_{k-2}} \mbf{Y}_{k-3} + \sqrt{\beta_k \beta_{k-1} - \beta_k \beta_{k-1} \beta_{k-2}} \epsilon^{*}_{k-3} + \sqrt{1 - \beta_k \beta_{k-1}} \epsilon_{k-2} \\
                  & = \underbrace{\sqrt{\beta_k \beta_{k-1} \beta_{k-2}} \mbf{Y}_{k-3} + {\color{blue} \sqrt{1 - \beta_k \beta_{k-1} \beta_{k-2}}} \epsilon_{k-3}}_{\text{recursion}}                            \\
                  & = \dots                                                                                                                                                                                            \\
                  & = \sqrt{\prod_{i=1}^{k} \beta_i} \mbf{Y}_0 + \sqrt{1-\prod_{i=1}^{k} \beta_i} \epsilon_0       \\
                  & = \sqrt{\alpha_k} \mbf{Y}_0 + \sqrt{1 - \alpha_k} \epsilon_0
    \end{aligned}
$}
\end{equation}
with $\epsilon_{i}, \epsilon_{i}^{*}$ for all $i = 0, \dots, k$ are i.i.d samples from $\mca{N}(0, \mbf{I})$.
In this way, we derive $q(\mbf{Y}_K | \mbf{Y}_0, \mbf{X})$ as in the form of Eq.~\ref{eq: ddqpm_forward_final}.

\noindent \textbf{Gaussianity of Markovian reverse process}.
Training the approximate denoising transition, i.e., {\it trajectory generator}, $p_{\theta}(\mbf{Y}_{k-1} | \mbf{Y}_{k}, \mbf{X})$ is to minimize its distance from $q(\mbf{Y}_{k-1} | \mbf{Y}_{k}, \mbf{X})$, which, however, is generally intractable.
A common way to mitigate such intractability is by conditioning on $\mbf{Y}_0$ additionally, with the Bayes' Rule being applied as
\begin{equation}
\resizebox{0.9\linewidth}{!}{$
    \begin{aligned}
        q(\mbf{Y}_{k-1} | \mbf{Y}_{k}, \mbf{Y}_0, \mbf{X}) & = \frac{q(\mbf{Y}_{k} | \mbf{Y}_{k-1}, \mbf{Y}_0, \mbf{X}) q(\mbf{Y}_{k-1} | \mbf{Y}_0, \mbf{X})}{q(\mbf{Y}_{k} | \mbf{Y}_0, \mbf{X})}  \\
                                                           & = \frac{{\color{red} q(\mbf{Y}_{k} | \mbf{Y}_{k-1}, \mbf{X})} {\color{blue} q(\mbf{Y}_{k-1} | \mbf{Y}_0, \mbf{X})} }{{\color{orange} q(\mbf{Y}_{k} | \mbf{Y}_0, \mbf{X})}} \\
                                                           & =\frac{ {\color{red} \mathcal{N}\left(\mathbf{Y}_k ; \sqrt{\beta_k} \mathbf{Y}_{k-1},\left(1-\beta_k\right) \mathbf{I}\right)} {\color{blue} \mathcal{N}\left(\mathbf{Y}_{k-1} ; \sqrt{\alpha_{k-1}} \mathbf{Y}_0,\left(1-\alpha_{k-1}\right) \mathbf{I}\right)}}{{\color{orange} \mathcal{N}\left(\mathbf{Y}_k ; \sqrt{\alpha_k} \mathbf{Y}_0,\left(1-\alpha_k\right) \mathbf{I}\right)}}                               \\
         & \propto \exp \left\{-\left[\frac{\left(\mathbf{Y}_k-\sqrt{\beta_k} \mathbf{Y}_{k-1}\right)^2}{2\left(1-\beta_k\right)}+\frac{\left(\mathbf{Y}_{k-1}-\sqrt{\alpha_{k-1}} \mathbf{Y}_0\right)^2}{2\left(1-\alpha_{k-1}\right)}-\frac{\left(\mathbf{Y}_k-\sqrt{\alpha_k} \mathbf{Y}_0\right)^2}{2\left(1-\alpha_k\right)}\right]\right\}                                     \\
         & =\exp \left\{-\frac{1}{2}\left[\frac{\left(\mathbf{Y}_k-\sqrt{\beta_k} \mathbf{Y}_{k-1}\right)^2}{1-\beta_k}+\frac{\left(\mathbf{Y}_{k-1}-\sqrt{\alpha_{k-1}} \mathbf{Y}_0\right)^2}{1-\alpha_{k-1}}-\frac{\left(\mathbf{Y}_k-\sqrt{\alpha_k} \mathbf{Y}_0\right)^2}{1-\alpha_k}\right]\right\}                                                                           \\
         & =\exp \left\{-\frac{1}{2}\left[\frac{\left(-2 \sqrt{\beta_k} \mathbf{Y}_k \mathbf{Y}_{k-1}+\beta_k \mathbf{Y}_{k-1}^2\right)}{1-\beta_k}+\frac{\left(\mathbf{Y}_{k-1}^2-2 \sqrt{\alpha_{k-1}} \mathbf{Y}_{k-1} \mathbf{Y}_0\right)}{1-\alpha_{k-1}}+C\left(\mathbf{Y}_k, \mathbf{Y}_0\right)\right]\right\}                                                               \\
         & \propto \exp \left\{-\frac{1}{2}\left[-\frac{2 \sqrt{\beta_k} \mathbf{Y}_k \mathbf{Y}_{k-1}}{1-\beta_k}+\frac{\beta_k \mathbf{Y}_{k-1}^2}{1-\beta_k}+\frac{\mathbf{Y}_{k-1}^2}{1-\alpha_{k-1}}-\frac{2 \sqrt{\alpha_{k-1}} \mathbf{Y}_{k-1} \mathbf{Y}_0}{1-\alpha_{k-1}}\right]\right\}                                                                                  \\
         & =\exp \left\{-\frac{1}{2}\left[\left(\frac{\beta_k}{1-\beta_k}+\frac{1}{1-\alpha_{k-1}}\right) \mathbf{Y}_{k-1}^2-2\left(\frac{\sqrt{\beta_k} \mathbf{Y}_k}{1-\beta_k}+\frac{\sqrt{\alpha_{k-1}} \mathbf{Y}_0}{1-\alpha_{k-1}}\right) \mathbf{Y}_{k-1} \right]\right\}                                                                                                    \\                                           
         & =\exp \left\{-\frac{1}{2}\left[\frac{1-\alpha_k}{\left(1-\beta_k\right)\left(1-\alpha_{k-1}\right)} \mathbf{Y}_{k-1}^2-2\left(\frac{\sqrt{\beta_k} \mathbf{Y}_k}{1-\beta_k}+\frac{\sqrt{\alpha_{k-1}} \mathbf{Y}_0}{1-\alpha_{k-1}}\right) \mathbf{Y}_{k-1}\right]\right\}                                                                                                \\
         & =\exp \left\{-\frac{1}{2}\left(\frac{1-\alpha_k}{\left(1-\beta_k\right)\left(1-\alpha_{k-1}\right)}\right)\left[\mathbf{Y}_{k-1}^2-2 \frac{\left(\frac{\sqrt{\beta_k} x_k}{1-\beta_k}+\frac{\sqrt{\alpha_{k-1}} x_0}{1-\alpha_{k-1}}\right)}{\frac{1-\alpha_k}{\left(1-\beta_k\right)\left(1-\alpha_{k-1}\right)}} \mathbf{Y}_{k-1}\right]\right\}                        \\
         & =\exp \left\{-\frac{1}{2}\left(\frac{1-\alpha_k}{\left(1-\beta_k\right)\left(1-\alpha_{k-1}\right)}\right)\left[\mathbf{Y}_{k-1}^2-2 \frac{\left(\frac{\sqrt{\beta_k} \mathbf{Y}_k}{1-\beta_k}+\frac{\sqrt{\alpha_{k-1}} \mathbf{Y}_0}{1-\alpha_{k-1}}\right)\left(1-\beta_k\right)\left(1-\alpha_{k-1}\right)}{1-\alpha_k} \mathbf{Y}_{k-1}\right]\right\}               \\
         & =\exp \left\{-\frac{1}{2}\left(\frac{1}{\frac{\left(1-\beta_k\right)\left(1-\alpha_{k-1}\right)}{1-\alpha_k}}\right)\left[\mathbf{Y}_{k-1}^2-2 \frac{\sqrt{\beta_k}\left(1-\alpha_{k-1}\right) \mathbf{Y}_k+\sqrt{\alpha_{k-1}}\left(1-\beta_k\right) \mathbf{Y}_0}{1-\alpha_k} \mathbf{Y}_{k-1}\right]\right\}                                                           \\
         & \propto \mathcal{N}(\mathbf{Y}_{k-1} | \mathbf{X}; \underbrace{\frac{\sqrt{\beta_k}\left(1-\alpha_{k-1}\right) \mathbf{Y}_k+\sqrt{\alpha_{k-1}}\left(1-\beta_k\right) \mathbf{Y}_0}{1-\alpha_k}}_{\mu_q\left(\mathbf{Y}_k, \mathbf{Y}_0\right)}, \underbrace{\left.\frac{\left(1-\beta_k\right)\left(1-\alpha_{k-1}\right)}{1-\alpha_k} \mathbf{I}\right)}_{\boldsymbol{\Sigma}_q(k)},
    \end{aligned}
$}
\end{equation}
which is also a Gaussian in relation to the forward process.
Moreover, we train diffusion models $p_{\theta}(q(\mbf{Y}_{k-1} | \mbf{Y}_{k}, \mbf{X})$ to generate $\mbf{Y}_0$ from $\mbf{Y}_{K}$ conditioned on $\mbf{X}$.
Then replace $\mbf{Y}_0$ in $q(\mbf{Y}_{k-1} | \mbf{Y}_{k}, \mbf{Y}_0, \mbf{X})$ with the predictive mean $\mu_{\theta}(\mbf{Y}_{k})$.
This informs us the predictive model $p_{\theta}(q(\mbf{Y}_{k-1} | \mbf{Y}_{k}, \mbf{X})$ can also be parameterized as a Gaussian with the predictive means: $\mu_{\theta}(\mbf{Y}_k) = \frac{1}{\alpha_k^2} \left( \mbf{Y}_{k} - (1 - \alpha_k) \theta_k(\mbf{Y}_k) \right) $, leading to $\mbf{Y}_{k-1} \sim q(\mbf{Y}_{k-1} | \mbf{Y}_{k}, \mbf{X})$, such that
\begin{equation}
\resizebox{0.9\linewidth}{!}{$
    \begin{aligned}
        q(\mbf{Y}_{k-1} | \mbf{Y}_{k}, \mbf{X}) & \approx q(\mbf{Y}_{k-1} | \mbf{Y}_{k}, \mbf{Y}_0 = \mu_{\theta}(\mbf{Y}_{k}), \mbf{X}) \\ 
        & = \mathcal{N}(\mathbf{Y}_{k-1} | \mathbf{X}; \frac{\sqrt{\beta_k}\left(1-\alpha_{k-1}\right) \mathbf{Y}_k}{1-\alpha_k} + \frac{ \sqrt{\alpha_{k-1}}\left(1-\beta_k\right)}{1-\alpha_k} \cdot \frac{1}{\alpha_k^2} \left( \mbf{Y}_{k} - (1 - \alpha_k) \theta_t(\mbf{Y}_k) \right),  \left.\frac{\left(1-\beta_k\right)\left(1-\alpha_{k-1}\right)}{1-\alpha_k} \mathbf{I}\right) \\
        & = \mathcal{N}(\mathbf{Y}_{k-1} | \mathbf{X}; \frac{1}{\sqrt{\beta_k}} \left( \mbf{Y}_k - \frac{1 - \beta_k}{\sqrt{1 - \alpha_k}} \theta_{k}(\mbf{Y}_k) \right), \left.\frac{\left(1-\beta_k\right)\left(1-\alpha_{k-1}\right)}{1-\alpha_k} \mathbf{I}\right).
    \end{aligned}
$}
\end{equation}

\noindent \textbf{Derivation of Eq.~\ref{eq: ddpm_elbo_emp}}.
We have stated that the empirical objective of Eq.~\ref{eq: ddpm_elbo} is given by Eq.~\ref{eq: ddpm_elbo_emp}.
To approach this, we derive an equivalent decomposition of Eq.~\ref{eq: ddpm_elbo} as
\begin{equation}\label{eq: markov_true_reverse}
\resizebox{0.8\linewidth}{!}{$
    \begin{aligned}
        \max_{\theta} & \mbb{E}_{p(\mbf{Y}_{0}, \mbf{Y}_{1}, \dots, \mbf{Y}_{K} | \mbf{X})} \left[ \log \frac{ p_{\theta} (\mbf{Y}_{0:K} | \mbf{X}) }{ q(\mbf{Y}_{1:K} | \mbf{Y}_{0}, \mbf{X}) } \right] \\
        & = \underbrace{\mathbb{E}_{q(\mbf{Y}_{1} | \mbf{Y}_0, \mbf{X})} \left[ \log p_{\theta}(\mbf{Y}_0 | \mbf{Y}_1, \mbf{X}) \right]}_{L_1} - \underbrace{\mathbb{E}_{ q(\mbf{Y}_{K-1} | \mbf{Y}_0, \mbf{X}) } D_{\text{KL}}(q(\mbf{Y}_{K} | \mbf{Y}_{K-1}, \mbf{X}) || p(\mbf{Y}_{K}))}_{L_2} \\
        & \qquad \; \qquad  \; \qquad - \underbrace{\sum_{k=1}^{K-1} \mathbb{E}_{ q(\mbf{Y}_{k-1}, \mbf{Y}_{k+1} | \mbf{Y}_0, \mbf{X}) } D_{\text{KL}}(q(\mbf{Y}_{k} | \mbf{Y}_{k-1}, \mbf{X}) || p_{\theta}(\mbf{Y}_{k} | \mbf{Y}_{k+1}, \mbf{X}))}_{L_3},
    \end{aligned}
$}
\end{equation}
where $L_2$ has no trainable parameters and $L_1$ is overwhelmed by $L_3$ provided that $K$ is a large number.
Knowing that the Gaussian forms of both $q(\mbf{Y}_{k} | \mbf{Y}_{k-1}, \mbf{X})$ and $p_{\theta}(\mbf{Y}_{k} | \mbf{Y}_{k+1}, \mbf{X})$, we primarily focus on $L_3$ as each term is optimized by minimizing the KL-divergence between them.
Denote the mean value of true denoising transition by $\mu_q\left(\mathbf{Y}_k, \mathbf{Y}_0\right)$, we have
\begin{equation}
\resizebox{0.6\linewidth}{!}{$
    \begin{aligned}
     & \mu_q\left(\mbf{Y}_k, \mbf{Y}_0\right)=\frac{\sqrt{\beta_k}\left(1-\alpha_{k-1}\right) \mbf{Y}_k+\sqrt{\alpha_{k-1}}\left(1-\beta_k\right) \mbf{Y}_0}{1-\alpha_k}                                                                              \\
     & =\frac{\sqrt{\beta_k}\left(1-\alpha_{k-1}\right) \mbf{Y}_k+\sqrt{\alpha_{k-1}}\left(1-\beta_k\right) \frac{\mbf{Y}_k-\sqrt{1-\alpha_k} \epsilon_0}{\sqrt{\alpha_k}}}{1-\alpha_k}                                                                              \\
     & =\frac{\sqrt{\beta_k}\left(1-\alpha_{k-1}\right) \mbf{Y}_k+\left(1-\beta_k\right) \frac{\mbf{Y}_k-\sqrt{1-\alpha_k} \epsilon_0}{\sqrt{\beta_k}}}{1-\alpha_k}                                                                                                             \\
     & =\frac{\sqrt{\beta_k}\left(1-\alpha_{k-1}\right) \mbf{Y}_k}{1-\alpha_k}+\frac{\left(1-\beta_k\right) \mbf{Y}_k}{\left(1-\alpha_k\right) \sqrt{\beta_k}}-\frac{\left(1-\beta_k\right) \sqrt{1-\alpha_k} \epsilon_0}{\left(1-\alpha_k\right) \sqrt{\beta_k}} \\
     & =\left(\frac{\beta_k\left(1-\alpha_{k-1}\right)}{\left(1-\alpha_k\right) \sqrt{\beta_k}}+\frac{1-\beta_k}{\left(1-\alpha_k\right) \sqrt{\beta_k}}\right) \mbf{Y}_k-\frac{1-\beta_k}{\sqrt{1-\alpha_k} \sqrt{\beta_k}} \boldsymbol{\epsilon}_0                          \\
     & =\frac{1-\alpha_k}{\left(1-\alpha_k\right) \sqrt{\beta_k}} \mbf{Y}_k-\frac{1-\beta_k}{\sqrt{1-\alpha_k} \sqrt{\beta_k}} \boldsymbol{\epsilon}_0                                                                                                                                 \\
     & =\frac{1}{\sqrt{\beta_k}} \mbf{Y}_k-\frac{1-\beta_k}{\sqrt{1-\alpha_k} \sqrt{\beta_k}} \boldsymbol{\epsilon}_0.
    \end{aligned}
$}
\end{equation}
Accordingly, minimizing the KL-divergence can be equivalently approximated by reducing the distance between two mean values, (setting the equivalent standard deviation terms $\Sigma_q(k)$),
\begin{equation}\label{eq: consistency_1}
\resizebox{0.9\linewidth}{!}{$
    \begin{aligned}
        D_{\text{KL}} & (q(\mbf{Y}_{k} | \mbf{Y}_{k-1}, \mbf{X}) || p_{\theta}(\mbf{Y}_{k} | \mbf{Y}_{k+1}, \mbf{X})) \\
        & = D_{\text{KL}}(\mca{N}(\mbf{Y}_{k-1}|\mbf{X}; \mu_{\theta}, \Sigma_q(k)) || \mca{N}(\mbf{Y}_{k-1}|\mbf{X}; \mu_{\theta}, \Sigma_q(k))) \\
        & = \frac{1}{2 \sigma_q^{2}(k)} \left[ \lVert \frac{ \sqrt{\beta_k} ( 1 - \alpha_k) \mbf{Y}_k + \sqrt{\alpha_{k-1}}(1 - \beta_k) \theta_{k}(\mbf{Y}_k) }{ 1 - \alpha_k} - - \frac{\sqrt{\beta_k} ( 1 - \alpha_k) \mbf{Y}_k + \sqrt{\alpha_{k-1}}(1 - \beta_k) \mbf{Y}_0}{ 1 - \alpha_k}   \lVert_{2}^{2}\right] \\
        & = \frac{1}{2 \sigma_q^{2}(k)} \frac{ \alpha_{k-1} (1 - \beta_k)^2 }{ (1 - \alpha_k)^{2} } \left[ \lVert \theta_{k}(\mbf{Y}_k) - \mbf{Y}_0 \lVert_{2}^{2}\right],    \quad \text{with } \Sigma_q(k) = \sigma_q^{2}(k) \mbf{I}.
    \end{aligned}
$}
\end{equation}
Equivalently, Eq.~\ref{eq: consistency_1} can be interpreted as predicting the noise added to the input - as in Eq.~\ref{eq: ddpm_elbo_emp},
\begin{equation}\label{eq: consistency_2}
\resizebox{0.8\linewidth}{!}{$
    \begin{aligned}
        D_{\text{KL}} & (q(\mbf{Y}_{k} | \mbf{Y}_{k-1}, \mbf{X}) || p_{\theta}(\mbf{Y}_{k} | \mbf{Y}_{k+1}, \mbf{X})) \\
        & = D_{\text{KL}}(\mca{N}(\mbf{Y}_{k-1}|\mbf{X}; \mu_{\theta}, \Sigma_q(k)) || \mca{N}(\mbf{Y}_{k-1}|\mbf{X}; \mu_{\theta}, \Sigma_q(k))) \\
        & =\frac{1}{2 \sigma_q^2(k)}\left[\left\|\frac{1}{\sqrt{\beta_k}} \mbf{Y}_k-\frac{1-\beta_k}{\sqrt{1-\alpha_k} \sqrt{\beta_k}} \theta_{k}(\hat{\epsilon}_k) -\frac{1}{\sqrt{\beta_k}} \mbf{Y}_k+\frac{1-\beta_k}{\sqrt{1-\alpha_k} \sqrt{\beta_k}} \epsilon_0 \right\|_2^2\right] \\
             & =\frac{1}{2 \sigma_q^2(k)}\left[\left\|\frac{1-\beta_k}{\sqrt{1-\alpha_k} \sqrt{\beta_k}} \epsilon_0-\frac{1-\beta_k}{\sqrt{1-\alpha_k} \sqrt{\beta_k}} \theta_{k}(\hat{\epsilon}_k) \right\|_2^2\right]                                                                                       \\
             & =\frac{1}{2 \sigma_q^2(k)} \frac{\left(1-\beta_k\right)^2}{\left(1-\alpha_k\right) \beta_k}\left[\left\|\epsilon_0-\theta_{k}(\hat{\epsilon}_k) \right\|_2^2\right] ~.
    \end{aligned}
$}
\end{equation}
This means that Eq.~\ref{eq: consistency_1} and Eq.~\ref{eq: consistency_2} represent equivalent objectives but simply set the neural network-based {\it trajectory generator} to predict different targets.

\noindent \textbf{Derivation of non-Markovian reverse process Eq.~\ref{eq: non_mk_denoise}}.
Define an inference family with valid transition $q_{\gamma}(\mbf{Y}_{k-1} | \mbf{Y}_{k}, \mbf{X})$ stipulates a ``marginal'' match as
\begin{equation}
\resizebox{0.6\linewidth}{!}{$
    q_{\gamma}(\mbf{Y}_{k-1} | \mbf{Y}_{0}, \mbf{X}) = \int q_{\gamma} ( \mbf{Y}_{k-1} | \mbf{Y}_{k}, \mbf{Y}_{0}, \mbf{X}) q_{\gamma} (\mbf{Y}_k | \mbf{Y}_0, \mbf{X}) \deriv{\mbf{Y}_{k}}.
$}
\end{equation}
Without loss of generality, its posterior (i.e., true reverse transition), is applicable to a Gaussian as Eq.~\ref{eq: markov_true_reverse},
\begin{equation}
\resizebox{0.7\linewidth}{!}{$
    \begin{aligned}
        q_{\gamma} (\mbf{Y}_{k-1} & | \mbf{Y}_{k}, \mbf{X}) = \mca{N}(\mbf{Y}_{k-1} | \mbf{X}; a_k \mbf{Y}_k + b_k \mbf{Y}_0; \gamma_k^2 \mbf{I}) \\
        \mathrm{s.t.} \; & q_{\gamma}(\mbf{Y}_{k-1} | \mbf{Y}_0, \mbf{X}) = \int q_{\gamma}(\mbf{Y}_{k-1} | \mbf{Y}_k, \mbf{Y}_0, \mbf{X}) q_{\gamma}(\mbf{Y}_k | \mbf{Y}_0, \mbf{X}) \deriv{\mbf{Y}_k}, \\
        & q_{\gamma}( \mbf{Y}_{k} | \mbf{Y}_{0}, \mbf{X}) =\mca{N}(\mbf{Y}_{k} | \mbf{X}; \sqrt{\alpha_k} \mbf{Y}_0, (1 - \alpha_k) \mbf{I}) \\
        & \alpha_{k-1} = a_k \alpha_k + b_k, \quad 1 - \alpha_{k-1} = \sqrt{ (1 - \alpha_k) a_k^2 + \gamma_k^2} ~.
    \end{aligned}
$}
\end{equation}
This is solvable as
\begin{equation}
\resizebox{0.8\linewidth}{!}{$
    \begin{aligned}
        q_{\gamma} ( \mbf{Y}_{k-1} | \mbf{Y}_{k}, \mbf{Y}_{0}, \mbf{X}) & =  \mca{N}\left( \left( \alpha_{k-1} - \frac{ \sqrt{\alpha_k} \sqrt{ 1 - \alpha_{k-1} - \gamma_k^2
 }  }{ \sqrt{1 - \alpha_k} }  \right) \mbf{Y}_0 + \frac{ \sqrt{1 - \alpha_{k-1} - \gamma_k^2} }{ \sqrt{ 1 - \alpha_k} } \mbf{Y}_k, \gamma_k^{2} \mbf{I} \right) \\ 
        & = \mca{N}\left( \sqrt{\alpha_{k-1}} \mbf{Y}_0 + \sqrt{1 - \alpha_{k-1} - \gamma_k^2} \frac{ \mbf{Y}_k - \sqrt{\alpha_k} \mbf{Y}_0 }{ \sqrt{ 1 - \alpha_k} }, \gamma_k^{2} \mbf{I} \right),
    \end{aligned}
$}
\end{equation}
where the only variable is $\gamma_k$.
Since UPDD involves stochasticity in the predictive distribution, we simply set $\gamma_k = 0$ for all $k > 0$, the variance term of transition becomes 0, i.e., it is deterministic.
However, this is infeasible for VDM approaches (e.g., MID~\cite{gu2022stochastic}) because they have nothing stochastic to account for multi-modal trajectory generation but only the noise terms $\epsilon_k$ for all $k \geq 0$.

\noindent \textbf{Gaussian density of log-likelihood objective Eq.~\ref{eq: loss_llh}}.
Recall that we assume the bi-variate Gaussian density function for the prediction of each pedestrian at time $t$ for all $t > T$, sufficiently defined by predictive means $\mu_1, \mu_2$ and standard deviations $\sigma_1, \sigma_2$ for both coordinate axes, and their correlation coefficient $\rho$.
Denote the true coordinates of pedestrian $n$ at time $t$ by $d_{t,1}^{(n)}, d_{t,2}^{(n)}$, the negative log-likelihood over all pedestrians and future timesteps are thus calculated by,
\begin{equation}
\resizebox{0.6\linewidth}{!}{$
\begin{aligned}
        \mca{L}_{\text{llh}} = - \sum_{n=1}^{N} \sum_{t=T+1}^{T^{\prime}} \log \left\{ (2\pi)^{-n/2} \det \left( \mbf{\Sigma}_{t, (n)} \right)^{1/2} \exp \left( -\frac{1}{2} \mbf{\Upsilon}^{\top} \mbf{\Sigma}_{t, (n)}^{-1} \mbf{\Upsilon} \right)  \right\},
\end{aligned}
$}
\end{equation}
with
\begin{equation}
\resizebox{0.6\linewidth}{!}{$
    \mbf{\Upsilon} = 
        \begin{bmatrix} 
            d_{t,1}^{(n)} - \mu_{t, 1}^{n} \\ 
            d_{t,2}^{(n)} - \mu_{t, 2}^{n}
        \end{bmatrix}
    \qquad
    \mbf{\Sigma_{t, (n)}} = 
        \begin{bmatrix}
            \left(\sigma_{t, 1}^{(n)}\right)^{2} & \rho_{t}^{(n)} \sigma_{t, 1}^{(n)} \sigma_{t, 2}^{(n)} \\
            \rho_{t}^{(n)} \sigma_{t, 1}^{(n)} \sigma_{t, 2}^{(n)} & \left(\sigma_{t, 2}^{(n)}\right)^{2}
        \end{bmatrix}.
$}
\end{equation}

\subsubsection{Details of Network Architectures}
\label{sec: arch_details}
We propose a number of new neural architectures along with the UPDD framework.
Generally, there is less complexity in their structures than those used in recent studies, e.g., $40\%$ number of parameters than MID~\cite{gu2022stochastic}. More detailed information is provided in Tab.~\ref{tab: model_backbone}.
That being said, we {\it do not} consider specific model designs as a major contribution to this work, but look into how neural architectures and generative frameworks interact in terms of complexity, i.e., whether more powerful generative frameworks can lift the burden of neural architectural complexity.
We discuss the implementation details of these architectures for completeness.
Again, they can be seamlessly replaced by any other function approximators.

\begin{figure}[t]
    \centering
    \includegraphics[width=0.9\linewidth]{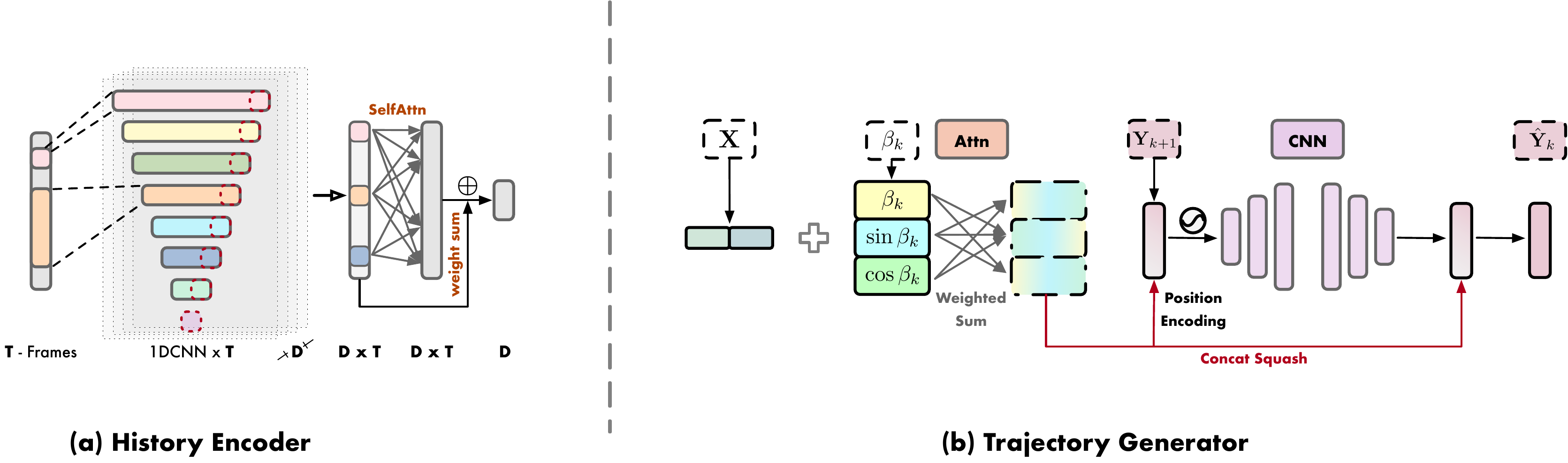}
    \caption{Illustrative demonstrations of {\bf Left}: the history encoder $\phi(\cdot)$~(as well as part of neighbor encoder $\psi(\cdot)$ implemented with paralleled single-layered CNNs and self-attention. {\bf Right}: how guidance information $\mbf{X}$ is conditioned to the reverse diffusion in each step.}
    \label{fig: supp_network}
\end{figure}

\noindent \textbf{History encoder}.
The historical path of each pedestrian is described by consecutive 2D coordinates in $T$ timesteps.
Resembling common practices, we also compute velocity and acceleration from successive steps.
Denote these input features of $N$ pedestrians by $\mbf{f} \in \mbb{R}^{N \times T \times 6}$, we then have $T$ 1D-CNNs with each having $D$ kernels applied to them as:
\begin{equation}
\resizebox{0.8\linewidth}{!}{$
\begin{aligned}
    \mbf{X}_{\text{hist}} = \mathrm{SelfAttn} \left( \big \lVert_{t=1}^{T} \left\{ \mathrm{CNN}_{t}(\mbf{f})[..., -1] \right\} \right) \in \mbb{R}^{N \times D}, \quad \text{with } k=t, p=0, s=1.
\end{aligned}
$}
\end{equation}
where $\big\lVert$ denotes vector concatenation. 
Each 1D-CNN has $D$ equal-size kernels, with kernel size $k$ being $t$ iterating from $1$ to $T$, padding size is $0$, and stride is $1$.
This leads to $D \times k$ feature maps for each CNN, having decreasing sizes as $k$ increases.
We concatenate the last step (which has the most implications for the $T+1$ step) of each 1D-CNN's output and apply a self-attention to obtain the reweighted sum of concatenated output features, resulting in the historical features $\mbf{X}_{\text{hist}}$ (depicted in Fig.~\ref{fig: supp_network}).

\noindent \textbf{Neighbor encoder}.
Consider any pedestrian $\forall \, n \in [1, N]$ to be the main subject of observation. 
The proximity of pedestrians to the main subject provides information on their social interactions.
The neighboring information is derived by the history information as well.
We weigh and aggregate the neighboring effects based on the inverse of Euclidean space between other pedestrians $j \neq n, j \in [1, N]$ and $n$.
The aggregated neighboring information forms unified interaction features of all neighbors in the past.
Following, the aggregated features are then passed into the history encoder~(with the same structure as $\phi(\cdot)$ but not sharing the parameters) to obtain $\mbf{X}_{\text{nei}}$,
\begin{equation}
\resizebox{0.8\linewidth}{!}{$
    \begin{aligned}
        \mbf{X}_{\text{nei}} & = \mathrm{SelfAttn} \left( \big \lVert_{t=1}^{T} \left\{ \mathrm{CNN}_{t}(\bar{\mbf{f}})[..., -1] \right\} \right) \in \mbb{R}^{N \times D}, \quad \text{with } k=t, p=0, s=1.  \\
        \text{with} \; \bar{\mbf{f}} & = \sum_{j \in N} \lambda_j \mbf{f}_j, \; \mbf{f}_j \in \mbb{R}^{N \times T \times 6}, \; \lambda_j \propto 1 / d(n, j), \quad \text{for all } n \in [1, N].
    \end{aligned}
$}
\end{equation}

\noindent \textbf{Distribution converter}.
It takes the 2D exact locations as input and output the 5D vectors composed of $[\mu_1, \mu_2, \sigma_1, \sigma_2, \rho]$, implemented as a single layer 1D CNN with the number of output feature maps as $5$, and $k=1, p=0, s=1$ referring to kernel size, padding and stride, respectively.

\noindent \textbf{Trajectory generator}.
In Sec.~\ref{sec: diffusion}, we state that the trajectory generator $\theta$ is a generative model conditioned on the guidance information $\mbf{X}$.
We first concatenate the history information and neighbor information $\mbf{X} := [\mbf{X}_{\text{hist}}, \mbf{X}_{\text{nei}}]$.
Since the guidance is applied to each diffusion step $k$ throughout the diffusion, we do the following the merge $\mbf{X}$ with forward coefficient $\beta_k$ (also depicted in Fig.~\ref{fig: supp_network}).
We first broadcast $\mbf{X}$ to add it with $\bar{\beta}_k := [\beta_k, \sin \beta_k, \cos \beta_k]$ derived from $\beta_k$.
Then, we apply self-attention to adaptively discover which component of $\bar{\beta}_k$ contributes the most to $\mbf{X}$.
This is followed by a {ConcatSquashLinear}\footnote[7]{we follow the implementation in \url{https://github.com/rtqichen/ffjord/blob/master/lib/layers/diffeq_layers/basic.py}} to merge the guide information with $\mbf{Y}$, i.e., sufficient statistics of future trajectory, and a positional encoding to further takes into account the temporal dependencies within data itself.
After that, the conditioned input is passed through $T^{\prime} / 2$ single layer 1D-CNNs with different feature maps, kernel sizes and paddings, ensuring the alignment of feature dimensions~(see Sec.~\ref{sec: supp_hparams} for detailed configuration).
Lastly, the concatenated output feature maps (they are in the same shape except for the number of kernels) go through three ConcatSquashLinear layers to obtain the targeted prediction for $\hat{Y}_{k}$ with re-weighted $\bar{\beta}_k$.

\section{Experiments and Analysis}

\subsection{Experiment Setup}

\subsubsection{Datasets}
Our empirical evaluations are based on the two real-world pedestrian trajectory benchmarks: ETH/UCY dataset and Standford Drone~(SDD) dataset.
ETH/UCY dataset consists of five sub-datasets, ETH, HOTEL, UNIV, ZARA1, ZARA2, whereas SDD dataset has 20 scenes.

\subsubsection{Baselines}

We compare UPDD with a range of state-of-the-art, including deterministic models~(D), goal-conditioned models~(G), implicit models~(I) and explicit models~(E).
This is a brief overview of these baselines.
\begin{itemize}
    \item Social-LSTM~\cite{alahi2016social}~(D) incorporates social-pooling operations and LSTM layers to predict the future trajectories of multiple agents in a social scene, taking into account both the spatial and temporal dependencies among the agents.
    \item STT~\cite{monti2022many}~(D) presents a consistency-based scheme to identify and localize anomaly movements in a video sequence. STT adopts a teacher-student framework where the teacher predicts the future frames while the student detects inconsistencies between the predicted and actual frames, which corresponds to anomalous events.
    \item Social-GAN~\cite{gupta2018social}~(I) extends Social-LSTM to generate the plausible trajectories with a Generative Adversarial Net~(GAN), where a generator network is trained to generate trajectories that implicitly match the statistical distribution of ground-truth, and a discriminator network is trained to distinguish between real and generated trajectories.
    \item SoPhie~\cite{sadeghian2019sophie}~(I) follows Social GAN, but uses an Attentive GAN and models scene context as auxiliary information.
    \item Social-BIGAT~\cite{kosaraju2019social}~(I) constructs a social interaction graph with Graph Attention Networks, and learns a reversible transformation between each scene and its latent noise vector by using BiCycle-GAN.
    \item MG-GAN~\cite{dendorfer2021mg}~(I) leverages multiple generators to prevent the model from generating out-of-distribution trajectories.
    \item Trajectron++~\cite{salzmann2020trajectron++}~(E) focuses on modeling the social interactions in terms of both spatial and temporal dependencies between pedestrians.
    \item Social-STGCNN~\cite{mohamed2020social}~(E) combines a graph convolutional neural network and a temporal CNN to predict the multiple future steps in a single shot.
    \item Causal-STGCNN~\cite{chen2021human}~(E) additionally incorporates counterfactual data augmentation into Social-STGCNN by perturbing ground-truth trajectories and learning to select the trajectory with the highest likelihood all mixture of trajectories.
    \item DMRGCN~\cite{bae2021disentangled}~(E) disentangle individual behaviors from their social interactions to improve the accuracy and interpretability of predictions.
    \item PECNet~\cite{mangalam2020not}~(G) is a pioneering work on goal-conditional methods, which first estimate the pedestrian's goal and then predict the trajectory conditional on the goal. 
    \item MemoNet~\cite{MemoNet}~(G) introduces a retrospective memory mechanism to enhance the performance of the model in this view.
    \item MID~\cite{gu2022stochastic}~(I) is a recent attempt to combine diffusion models with trajectory prediction. MID approximates reverse diffusion processes to gradually remove predictive indeterminacy and eventually obtain an exact predictive trajectory for each pedestrian.
    \item Leapfrog~\cite{Leapfrog}~(I) uses DDIM to shorten the diffusion chain and sample coarse trajectories and random noises to ensure multi-modality. Still, diffusion is applied on exact locations and executed once for each trajectory, which does not actually reduce the overall count of reverse diffusion processes.

\end{itemize}

\subsubsection{Metrics}
We report Minimum Average Displacement Error~(minADE) and Minimum Final Displacement Error~(minFDE) for quantitative evaluation.
ADE computes the average predictive error between the best estimation and the ground-truth trajectory; FDE measures the displacement error at the last point.
We follow the common evaluation criteria as in most studies~\cite{salzmann2020trajectron++, gu2022stochastic}.

\subsubsection{Hyperparameters}
\label{sec: supp_hparams}
Both $\phi(\cdot)$ and $\psi(\cdot)$ are with 8 parallel 1D-CNNs, $\theta(\cdot)$ is with 6 parallel 1D-CNNs.
We set the dimension $d=128$ for all hidden layers/embeddings/etc. 

The configuration of CNNs in $\phi$ and $\psi$ are $k=1, \dots, T, s=1, p=0$ with $d=128$ feature maps.
The self-attention is implemented with $d=128$.

As for $\theta$, the configuration of CNNs in  are $k=1, 3, 5, 7, 9, 11$, padding $p=0, 1, 2, 3, 4, 5$, stride $s=1$, with $d=128, 64, 16, 16, 16, 16$ feature maps.
The four ConcatSquashLinear layers are with input sizes of $5, 256, 128, 64$, and output sizes of $256, 128, 64, 5$.

We set the batch size to 256, using Adam optimizer with an initial learning rate of 0.001 and the Exponential scheduler with a multiplicative factor of 0.95.
The coefficients for three objectives are $\lambda_{\text{llh}} = 10^{-5}, \lambda_{\text{emp}} = 10, \lambda_{\text{con}} = 10$ for synchronized optimization in similar orders of magnitude, respectively.
$\alpha_K$ is a sequence uniformly descending sampled from the interval (0, 1], where $K$ is the total number of diffusion steps.
The training epochs for UPDD are set to 100, whereas the approaches for comparison are set by following either the official implementation or the descriptions in the papers.

\begin{table}[ht]
    \centering
    \resizebox{\linewidth}{!}{%
        \begin{tabular}{l|ccccc|c}
            \toprule
            \textbf{Method}             & \textbf{ETH} & \textbf{HOTEL} & \textbf{UNIV} & \textbf{ZARA1} & \textbf{ZARA2} & \textbf{AVG}       \\
            \midrule
            Social-LSTM~\cite{alahi2016social}                  & 1.09/2.94    & 0.86/1.91      & 0.61/1.31     & 0.41/0.88      & 0.52/1.11      & 0.70/1.52          \\
            STT~\cite{monti2022many}                         & 0.54/1.10    & 0.24/0.46      & 0.57/1.15     & 0.45/0.94      & 0.36/0.77      & 0.43/0.88          \\
            \midrule
            Social-GAN~\cite{gupta2018social}                   & 0.81/1.52    & 0.72/1.61      & 0.60/1.26     & 0.34/0.69      & 0.42/0.84      & 0.58/1.18          \\
            SoPhie~$^{\dagger}$~\cite{sadeghian2019sophie}         & 0.70/1.43    & 0.76/1.67      & 0.54/1.24     & 0.30/0.63      & 0.38/0.78      & 0.54/1.15          \\
            Social-BIGAT~$^{\dagger}$~\cite{kosaraju2019social}   & 0.69/1.29    & 0.49/1.01      & 0.55/1.32     & 0.30/0.62      & 0.36/0.75      & 0.48/1.00          \\
            MG-GAN~$^{\dagger}$~\cite{dendorfer2021mg}         & 0.47/0.91    & 0.14/0.24      & 0.54/1.07     & 0.36/0.73      & 0.29/0.60      & 0.36/0.71          \\
            \midrule
            Trajectron++$^*$~\cite{salzmann2020trajectron++}            & 0.67/1.18    & 0.18/0.28      & 0.30/0.54     & 0.25/0.41      & 0.18/0.32      & 0.32/0.55          \\
            Social-STGCNN~\cite{mohamed2020social}                    & 0.64/1.11    & 0.49/0.85      & 0.44/0.79     & 0.34/0.53      & 0.30/0.48      & 0.44/0.75          \\
            Causal-STGCNN~\cite{chen2021human}                    & 0.64/1.00    & 0.38/0.45      & 0.49/0.81     & 0.34/0.53      & 0.32/0.49      & 0.43/0.66          \\
            DMRGCN~\cite{bae2021disentangled}                      & 0.60/1.09    & 0.21/0.30      & 0.35/0.63     & 0.29/0.47      & 0.25/0.41      & 0.34/0.58          \\
            \midrule
            PECNet$^{\top}$~\cite{mangalam2020not}             & 0.54/0.87    & 0.18/0.24      & 0.35/0.60     & 0.22/0.39      & 0.17/0.30      & 0.29/0.48          \\
            MemoNet$^{\top}$~\cite{MemoNet}                         & 0.40/0.61    & \textbf{0.11}/\textbf{0.17}      & 0.24/0.43     & \underline{0.18}/0.32      &  \underline{0.14}/\underline{0.24}      & 0.21/0.35          \\
            \midrule
            MID$^*$~\cite{gu2022stochastic}        & 0.54/0.82    & 0.20/0.31      & 0.30/0.57     & 0.27/0.46      &  0.20/0.37      & 0.30/0.51 \\
            Leapfrog~\cite{Leapfrog}                      & 0.39/0.58    & \textbf{0.11}/\textbf{0.17}      & 0.26/0.43     & \underline{0.18}/\textbf{0.26}     &  \textbf{0.13}/\textbf{0.22}     &  0.21/\underline{0.33} \\
            \midrule
            UPDD(10/100)[10+10]                & 0.40/0.82    & 0.22/0.43      &  0.24/0.49     & 0.31/0.64      & 0.25/0.50      &  0.28/0.58          \\
            UPDD(100/200)[10+10]               &  0.41/0.86    & 0.19/0.36     & 0.23/0.49    &  0.23/0.49      & 0.21/0.43      & 0.25/0.53          \\
            \midrule
            UPDD(10/100)[10/10]                & \textbf{0.22}/\textbf{0.42}    & 0.17/0.30      &  \textbf{0.14}/\textbf{0.28}     &  \textbf{0.16}/\underline{0.30}      &  \underline{0.14}/0.31      & \textbf{0.17}/\textbf{0.32}          \\
            UPDD(100/200)[10/10]               & \underline{0.30}/\underline{0.50}    & 0.18/0.30      &  \textbf{0.14}/\underline{0.29}     &  \underline{0.18}/0.33      & 0.17/0.30      & \underline{0.19}/0.34          \\
            \bottomrule
        \end{tabular}
    }
    \caption{
        minADE/minFDE results (meters) ETH/UCY dataset.
        $\dagger$ indicates models use auxiliary image information of scene context.
        $*$ indicates a correction of results caused by the leakage of future paths.
        ${\top}$ means goal-conditioned methods - first estimate the goal of the pedestrian and then predict the trajectory conditioned on the goal.
        Our UPDD (Step)[Sample] denotes different step sizes and sampling methods.
        We {\bf bolded} the best results, while \underline{underlining} the second best.
    }
    \label{tab: overall_comparison}
\end{table}

\subsection{Comparison with State-of-the-Art}
\label{sec: quantitative}

\subsubsection{Quantitative Evaluation}
Tab.~\ref{tab: overall_comparison} compares UPDD with state-of-the-art in terms of ADE and FDE.
We include two different diffusion steps with UPDD, in the form of $(S/K)$ indicating execution steps and lengths of Markovian diffusion.
Thanks to the explicit Gaussian density, UPDD does not require running reverse diffusion 20 times as VDMs to generate 20 trajectories. 
In variant $[10+10]$, we first run the reverse diffusion process 10 times and sample a diffusion outcome as the base.
We then sample 10 trajectories therefrom.
In variant $[10/10]$, reverse diffusion is run 10 times with 10 trajectories sampled from each outcome.
Observe that UPDD consistently demonstrates promising performances with ETH, UNIV, ZARA1 and AVG results.
Generated approaches are generally advantageous provided they output multi-modal predictions.
Moreover, both diffusion model-based approaches, Leapfrog and UPDD reports the lowest AVG errors among all approaches, showing improved expressiveness thanks to the flexible diffusion models.
In addition, MemoNet based on goal-conditioned models obtained excellent results on HOTEL. The reason may be that the trajectories of the HOTEL dataset are mostly straight, which helps the goal-conditioned based models.
UPDD, due to explicitly incorporating the self-uncertainty of pedestrians and extensively collecting social information, performs slightly worse in predicting trajectories that are nearly straight.
MID and Leapfrog predict future trajectory coordinates by directly encoding the trajectories themselves. Their randomness originates from the initial noise sampling, with each trajectory predicted independently and trajectories being mutually independent. Our UPDD, on the other hand, predicts sufficient statistics of the trajectory, resulting in a distribution of future trajectories from which we then sample. Our UPDD preserves the self-uncertainty of pedestrians, and the sampled trajectories belong to the predicted distribution, hence it may be slightly inferior to Leapfrog in scenarios with weak self-uncertainty. However, it's noteworthy that our UPDD does not require running a reverse diffusion process separately for each trajectory, and we compare efficiency in subsequent analyses.

Tab.~\ref{tab: ablation} also shows our exploration of how UPDD is affected by different sampling strategies.
We perform experiments on the datasets UNIV and SDD.
It can be seen that with almost no effect on time, our model can obtain more accurate predictions.
The variant [20/20], shows that our model increases in time complexity by just less than 2 times, but the accuracy is substantially improved.

\begin{table}[htbp]
\centering
\resizebox{0.6\linewidth}{!}{
\begin{tabular}{c|c|cc|cc}
\toprule
dataset               & method & step    & sample & ADE/FDE    & rel. time \\ \midrule
\multirow{9}{*}{UNIV} & MID    & 100     & 20     & 0.30/0.57  & 3.444     \\
                      & MID    & 100     & 100    & 0.21/0.32  & 14.717    \\
                      & MID    & 100     & 400    & 0.14/0.16  & 35.946    \\
                      & UPDD   & 10/100  & 10+10  & 0.24/0.49  & 1         \\
                      & UPDD   & 10/100  & 10/10  & 0.14/0.28  & 1.008     \\
                      & UPDD   & 10/100  & 20/20  & 0.07/0.15  & 1.152     \\
                      & UPDD   & 100/200 & 10+10  & 0.23/0.49  & 1.968     \\
                      & UPDD   & 100/200 & 10/10  & 0.14/0.29  & 2.05      \\
                      & UPDD   & 100/200 & 20/20  & 0.08/0.15  & 2.819     \\ \midrule
\multirow{4}{*}{SDD}  & MID    & 10      & 100    & 9.88/20.23 & 4.37      \\
                      & MID    & 100     & 100    & 4.80/9.88  & 25.2      \\
                      & UPDD   & 10/100  & 10/10  & 7.32/14.46 & 1         \\
                      & UPDD   & 100/200 & 10/10  & 6.59/13.90 & 2.73      \\ \bottomrule
\end{tabular}
}
\caption{
    Performance vs. diffusion step and sampling.
}
\label{tab: ablation}
\end{table}

\subsubsection{Qualitative Results}
\begin{figure}[htbp]
    \begin{center}
        \begin{small}
            \includegraphics[width=0.8\linewidth]{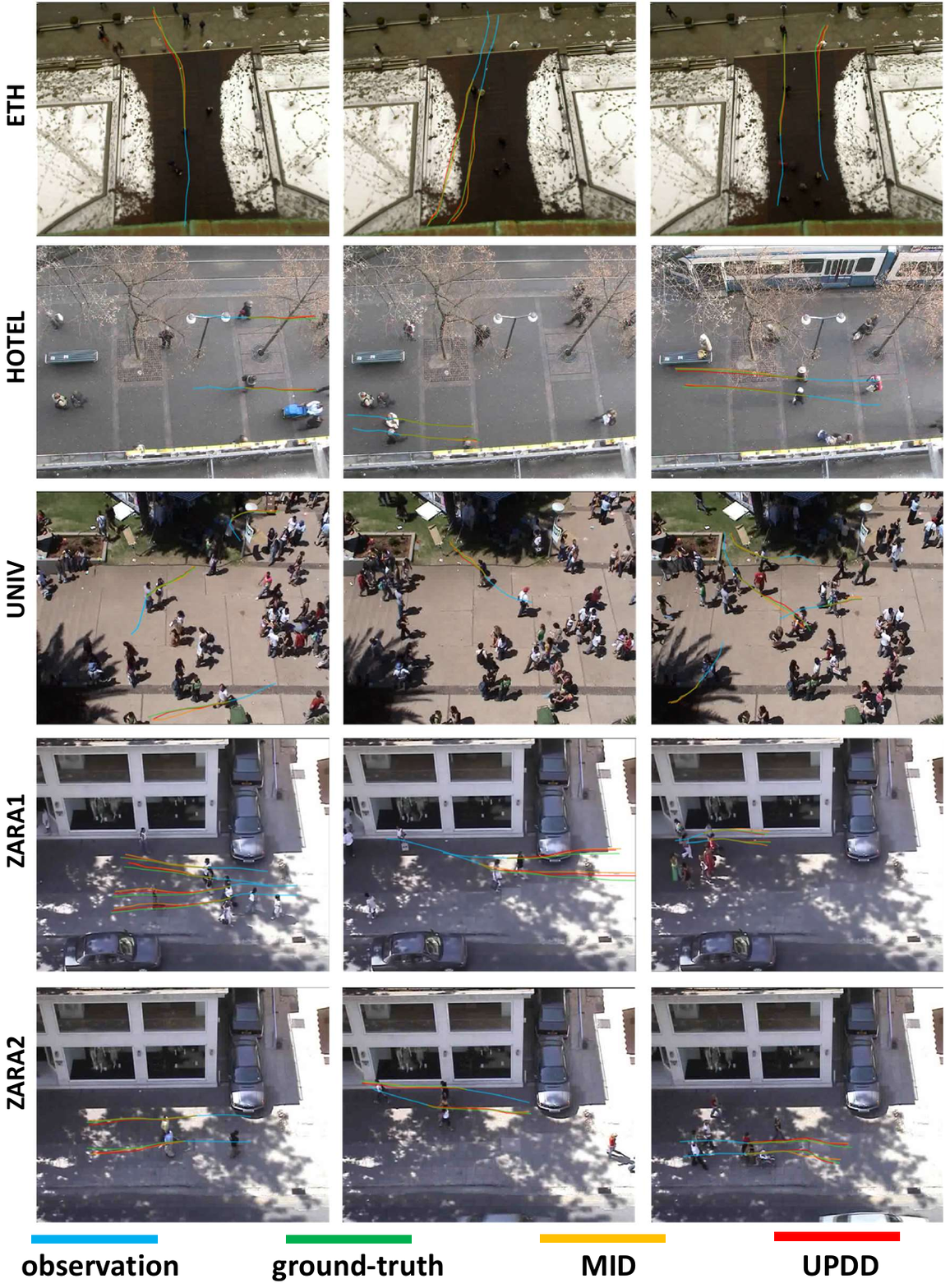}
        \end{small}
    \end{center}
    \caption{Qualitative Visualization}
    \label{fig: qualitative}
\end{figure}

We showcase visual comparisons of UPDD and MID with respect to the ground-truth trajectory as in Fig.~\ref{fig: qualitative}.

\begin{figure}[htbp]
    \centering
    \includegraphics[width=0.8\linewidth]{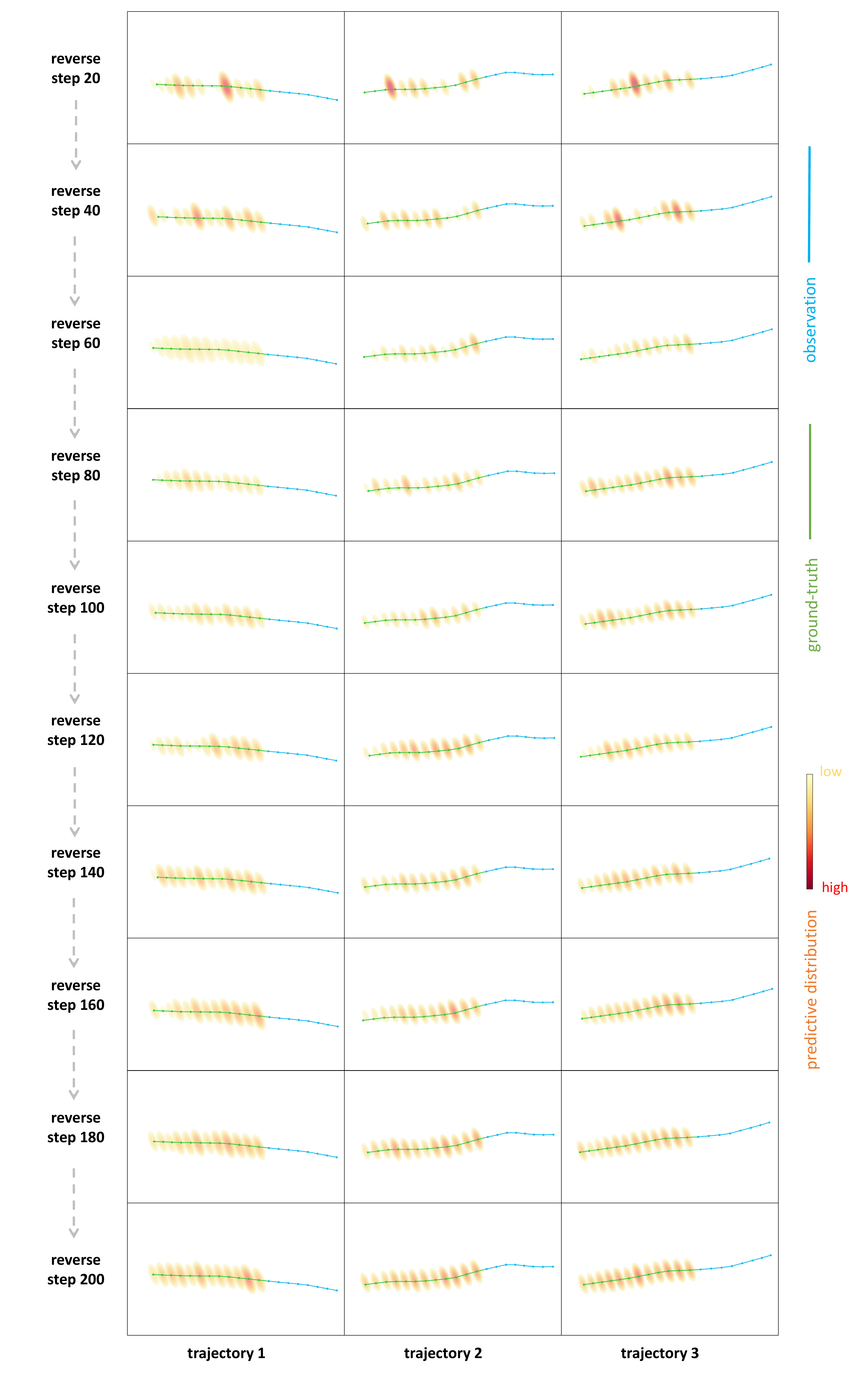}
    \caption{Visualizations of Predictive distribution of UPDD (100/200).}
    \label{fig: prob_vis}
\end{figure}

In addition, we provide visualizations of how the approximated reverse diffusion process evolves in steps.
We showcase UPDD (100/200) with execution steps $S = 100$ and length of diffusion chains $K = 200$ as in Fig.~\ref{fig: prob_vis}. 
The trajectories of three pedestrians are shown from left to right, while steps of reverse diffusion are shown from top to bottom.
UPDD predicts the upcoming locations based on blue-colored observed history through a gradient mustard-red predictive distribution for each location, whereas the ground truth future trajectory is shown in green.
As a visual aid, the gradient colors are drawn within a certain range with red marking a higher probability density.
Thus, some of the early diffusion steps are nearly invisible due to their low probability density.
We have observed that a well-trained UPDD~(100/200) forms the correct trend for most of the predictive distributions when K takes just 20 steps, despite scattered probability densities and shifted centroids.
Gradually, densities become concentrated and centroids approach ground truth as the reverse process continues.
The reverse diffusion model thus gradually fits the probability distribution of the trajectory as it proceeds.
Once we reach K = 200 to complete the reverse process, the predictive distribution already represents the future reasonably well.
Notably, the pedestrian's self-uncertainty can be seen in the predictive means being close to, but not completely overlapping the future trajectory.
Such uncertainty is identified and retained in the bi-variate Gaussian model predictions, allowing us to sample high-quality future trajectories therefrom.
Briefly, UPDD appreciates the expressiveness of diffusion models in exploring complex trajectory patterns whilst explicitly characterizing as well the inherent uncertainty and indeterminacy of the future.

\subsection{Ablation Studies}
\subsubsection{Efficiency}
Despite the immense potential for learning complex data distributions, diffusion models are notorious for their computational cost, often requiring hundreds or even thousands of reverse transitions to approximate the data distribution from standard Gaussian noise.
In order to address this significant limitation associated with diffusion model-based approaches, we introduce an effective workaround by adapting deterministically accelerated `diffusion'' models~\cite{song2020denoising} for trajectory prediction.
In terms of time complexity, we observe a remarkable improvement in relative time efficiency for UPDD, with roughly 3 to 4 times faster execution with just one iteration compared to MID, as illustrated in Tab.~\ref{tab: ablation}.
It's worth noting that while UPDD benefits from this speed boost, the same cannot be said for MID. Although both models utilize diffusion models, there is a fundamental difference in their approaches. UPDD parameterizes explicit density functions for future locations, whereas MID predicts exact coordinates. Consequently, MID relies entirely on sampled noises along with reverse diffusion to introduce predictive stochasticity for multi-modal generation, making it incompatible with deterministic acceleration.
On the other hand, UPDD can theoretically eliminate noise-induced randomness from each reverse transition, as the explicit predictive distribution naturally accommodates the self-uncertainty of pedestrians. Moreover, the time complexity can be further emphasized by the fact that reverse diffusion requires numerous iterations, such as 100 steps for MID. As a result, the complexity scales linearly with the actual execution steps, underscoring the advantages of shortening the reverse chain.

We have expanded our efficiency comparison between UPDD and MID, providing additional results. Tab.~\ref{tab: model_backbone} offers a comprehensive comparison of the model backbone in terms of time complexity and space complexity.
Under their respective standard models, UPDD exhibits a remarkable advantage with only $40\%$ of the number of parameters and 1/3 of the FLOPS compared to MID, while still achieving superior results.
Furthermore, we conducted a fair scaling down of the feature dimensions for both MID and UPDD proportionally. As expected, the model's performance diminishes as complexity is reduced. However, it's noteworthy that UPDD consistently outperforms MID in all metrics even when complexity is downgraded. Especially, under extreme conditions, we maintain good performance with very small resource consumption, and compare favorably with the advanced model in Tab.~\ref{tab: overall_comparison}.

In Tab. \ref{tab: additional}, we compare the results of MID (10)[20] and UPDD (10/100) [10/2] for the same 20 samples when the execution step is 10.
The results show that UPDD has an advantage when it comes to shortening the execution steps, which means that it is more appropriate in resource-limited scenarios.
Tab.~\ref{tab: additional} also shows our exploration of how UPDD is affected by different sampling strategies.
The first one variant [10+10] refers to the standard UPDD.
In variant [20/1], reverse diffusion is run 20 times with one trajectory sampled from each outcome, resembling MID.
In variant [1/20], a single run of reverse diffusion is conducted, but 20 trajectory samples are taken.
In this sense, variant [20/1] attributes all stochasticity to random noise throughout the diffusion, disregarding the self-uncertainty in terms of the Gaussian predictions.
Conversely, variant [1/20] may deteriorate due to a random sample given by the starting point of reverse diffusion (it only randomly samples a Gaussian noise once).
It appears that both samplings degrade model performance, suggesting that the diffusion model becomes more effective when combined with Gaussian predictive distributions.

\begin{table}[ht]

        \centering
        \resizebox{0.6\linewidth}{!}{%
            \begin{tabular}{c|c|cc}
            \toprule
            {\bf UNIV}             & {\bf \begin{tabular}[c]{@{}c@{}}Enc-dim/\\ Gen-dim\end{tabular}}     & {\bf ADE/FDE}      & {\bf \begin{tabular}[c]{@{}c@{}}FLOPs(b)/\\ params(m)\end{tabular}}   \\
            \midrule
                                   & 128/512* & 0.30/0.57 & 10.33/3.79      \\
                                   & 64/256   & 0.31/0.57 & 5.19/0.95  \\
                                   & 32/128   & 0.30/0.55 & 0.66/0.24  \\
                                  
            \multirow{-4}{*}{\begin{tabular}[c]{@{}c@{}}MID with \\ diffusion (100) \\ and sample (20)\end{tabular}}   & 16/64    & 0.32/0.59 & 0.17/0.06  \\
            \midrule
                                   & 128/256* & 0.24/0.49 & 3.27/1.48 \\
                                   & 64/128   & 0.23/0.49 & 0.83/0.37 \\
                                   & 32/64    & 0.26/0.54 & 0.21/0.09 \\
             \multirow{-4}{*}{\begin{tabular}[c]{@{}c@{}}UPDD with \\ diffusion (10/100)\\ and sample (10+10)\end{tabular}} & 16/32   & 0.27/0.55 & 0.05/0.02 \\
            \bottomrule
            \end{tabular}
        }
        \caption{Complexity comparison of model backbone variants on UNIV dataset. * is the original model. }
        \label{tab: model_backbone}

\end{table}

\begin{table}[ht]

        \centering
        \resizebox{0.5\linewidth}{!}{%
            \begin{tabular}{c|cc|c}
            \toprule
            {\bf UNIV}             & {\bf step}     & {\bf sample}      & {\bf ADE/FDE}   \\
            \midrule
             MID   & 10         & 20        & 0.42/0.72  \\
             UPDD  & 10/100     & 10/2     & 0.27/0.55  \\
             \midrule
             UPDD  & 100/200     & 10+10     & 0.23/0.49  \\
             UPDD  & 100/200     & 20/1    & 0.34/0.77  \\
             UPDD  & 100/200     & 1/20    & 0.41/0.79  \\

            \bottomrule
            \end{tabular}
        }
        \caption{Experiments on modelling diffusion step and sampling on UNIV dataset.}
        \label{tab: additional}

\end{table}

\subsubsection{Diffusion Steps}

We turn our attention to the impact of varying hyperparameters within the diffusion model, as outlined in Tab.~\ref{tab: ablation2}. 
Our analysis includes different variants with total diffusion steps of $K = 100, 200, 300$, all maintaining consistent execution steps at $S=100$. 
Our observations indicate that alterations in $K$ do not markedly influence the model's predictive capability, suggesting that the chain can be shortened with a modest ratio adjustment. 
Furthermore, we present a variant with $K=100$ and $S=10$, which significantly reduces generation time without compromising effectiveness. This setup could be favored when prioritizing efficiency.

Moreover, we experiment by substituting the deterministic acceleration in UPDD's diffusion model with the stochastic DDPM approach from MID, as per ~\cite{NEURIPS2020_4c5bcfec}, and refer to this variant as UPDD(P). 
In scenarios where $K(S)=10$, UPDD(P) might experience incomplete diffusion, resulting in a noticeable performance discrepancy compared to other models. 
However, when $K(S)$ is expanded to 100, UPDD(P)'s performance nearly aligns with that of UPDD(I). 
These findings imply that UPDD is inherently compatible with deterministic diffusion chains, as its stochastic nature is embedded within the predictive distribution, thus eliminating the dependence on noise sampling during the diffusion process.

\begin{table}[htbp]
\centering
\resizebox{0.8 \linewidth}{!}{
\begin{tabular}{c|c|ccc|cc}
\toprule
Variant & UPDD(I)   & UPDD(I)   & UPDD(I)   & UPDD(I)   & UPDD(P)   & UPDD(P)   \\ \midrule
Step   & 10/100    & 100/100   & 100/200   & 100/300   & 10        & 100       \\ \midrule
ADE/FDE & 0.24/0.49 & 0.25/0.49 & 0.23/0.49 & 0.23/0.50 & 0.36/0.66 & 0.26/0.51 \\ \bottomrule
\end{tabular}
}
\caption{
    Ablation with diffusion step on UNIV dataset. ``I'' refers to the DDIM scheme, which is our original model, and ``P'' refers to the DDPM scheme, which is the variant model. Step refers to ``execution steps'' and ``total diffusion steps''.
}
\label{tab: ablation2}
\end{table}

To sum up, based on the findings presented in Tab.~\ref{tab: ablation2}, UPDD offers greater efficiency advantages over VDMs from several key perspectives:
(1) UPDD employs more lightweight neural architectures.
(2) UPDD leverages the predictive stochasticity inherent in explicit distributions, enabling acceleration through deterministically shortened diffusion chains.
(3) The efficiency benefits become even more pronounced when generating a larger number of trajectories.

\subsubsection{Ablation with different Neural Architectures}

\begin{figure}[htbp]
    \centering
    \includegraphics[width=0.9\linewidth]{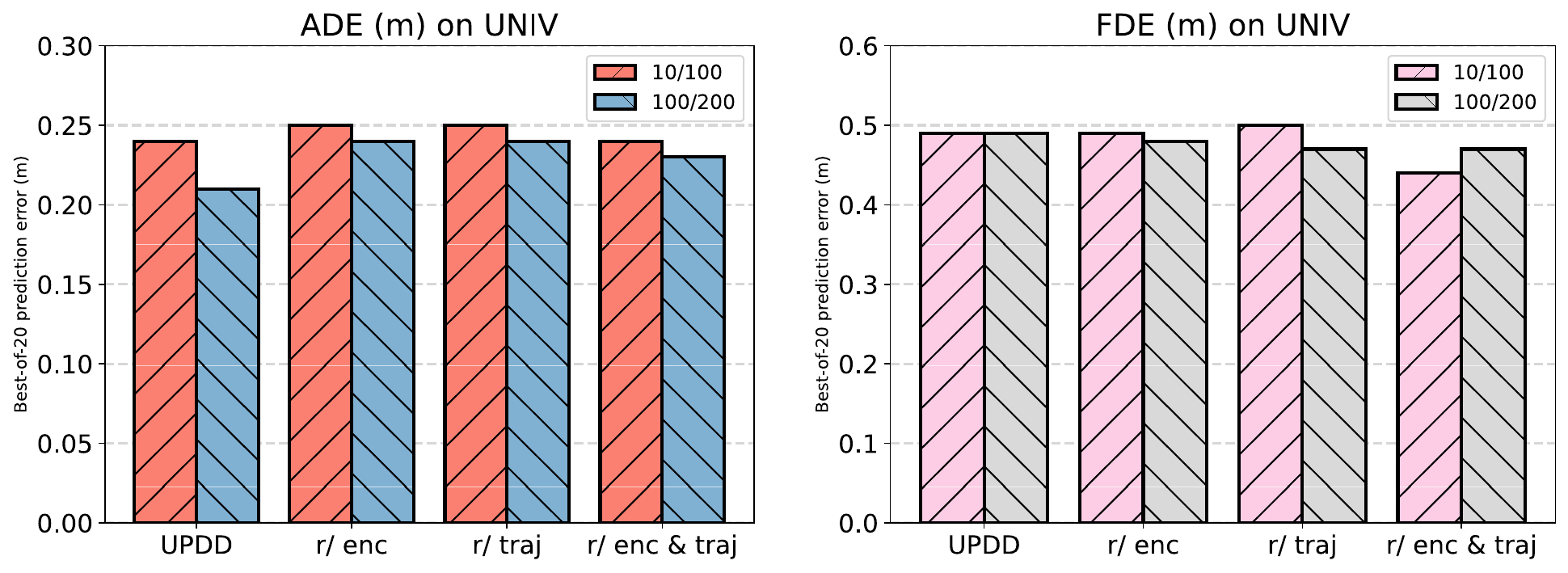}
    \caption{Additional ablation results with the proposed architectures. We report {\bf Left}: ADE {\bf Right}: FDE by {\it UPDD}: using the proposed architectures; {\it r/ enc}: replacing $\phi, \psi$ with the encoder used in MID~\cite{gu2022stochastic} and trajectron++~\cite{salzmann2020trajectron++}; {\it r/ traj}: replacing CNN-based $\theta$ with the Transformer-based trajectory generator~\cite{gu2022stochastic}; {\it r/ enc \& traj}: replacing $\phi, \psi, \theta$ with the complex ones used in~\cite{gu2022stochastic}.}
    \label{fig: supp_ablation_archs}
\end{figure}

\begin{figure}[htbp]
        \centering
        \includegraphics[width=0.5\linewidth]{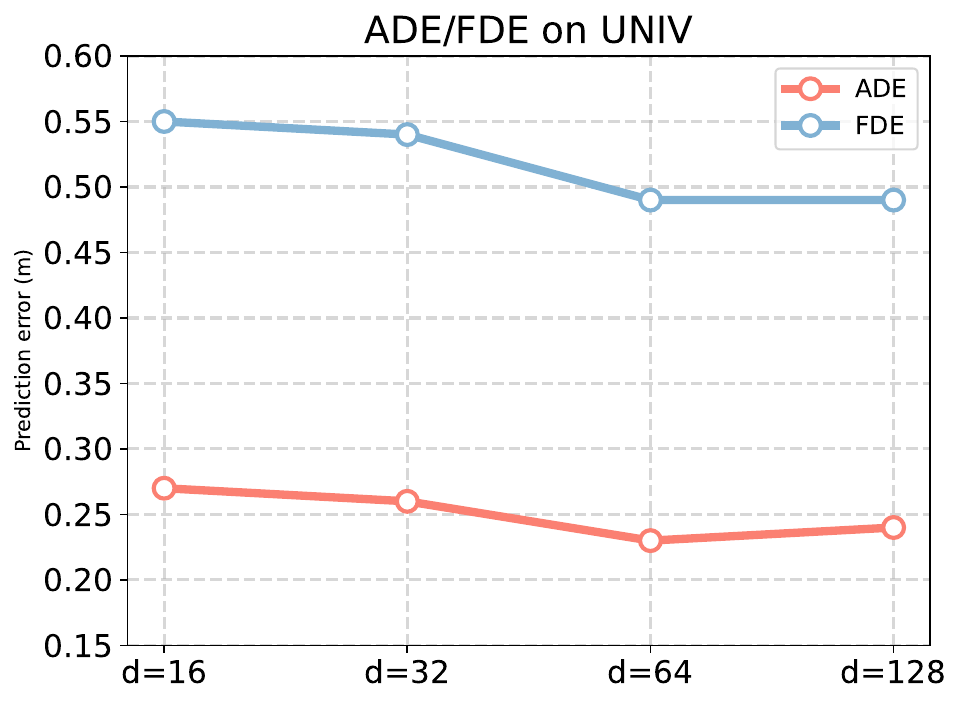}
        \caption{Ablation results of varying the values of $d$ with the proposed CNN-based architectures.}
        \label{fig: ablation_dimension}
\end{figure}

Following the details discussed in Sec.~\ref{sec: arch_details}, we conducted quantitative evaluations to assess the impact of the proposed CNN-based components, namely $\phi$, $\psi$, and $\theta$. The results are presented in Fig.~\ref{fig: supp_ablation_archs}.
For a direct comparison, we replaced $\phi$, $\psi$, and $\theta$ with their counterparts from MID~\cite{gu2022stochastic}, where the encoder is based on an LSTM-based spatial-temporal network, and the trajectory generator is constructed using a Transformer. Both dedicated architectures are considered to have higher capacity than the CNN-based ones discussed in this study, not only due to their larger number of trainable parameters. However, we did not observe significant performance impacts resulting from the use of neural architectures with greater potential capacity.

Additionally, in Fig.~\ref{fig: ablation_dimension}, we conducted ablation studies within the proposed CNN-based architectures by varying the hidden dimension $d$ within the range of 16 to 128. Generally, an increase in $d$ signifies higher model capacity. However, it's worth noting that the predictive performance did not improve proportionally with the increase in $d$.

In our assessment, this suggests that predictive performance seems to be more reliant on the generative framework's capabilities if it is already sufficiently powerful. We believe this warrants further investigation.

\section{Conclusion and Outlook}

We have presented an Uncertainty-aware Trajectory Prediction framework with distributional diffusion, which simulates the complex pedestrians' movements with a flexible diffusion model, and predicts uncertain future locations with explicit bi-variate Gaussian distributions, respectively.
While the distributional diffusion approximates the true denoising process that steadily improves confidence about future movement, the explicit predictive distributions of locations preserve inherent individual uncertainty that should not be eliminated.
By doing so, we can even bypass expensive sampling of diffusion models with deterministic accelerated means which previous models cannot.

\noindent \textbf{Outlook}.
UPDD requires careful choices for hybrid sampling with both diffusion and bi-variate Gaussian.
While sampling from the explicit Gaussian can accelerate the process, the quality of generations depends on the outcome of reverse diffusion.
It is worth exploring how to strike the right balance between efficiency and effectiveness.
At the same time, in the field of trajectory prediction, utilizing social information effectively while eliminating interference remains a current challenge. In the future, we plan to continue our efforts in effectively addressing trajectory prediction and more general sequence prediction problems using diffusion models. Our aim is to make optimal use of guiding information and enhance the efficiency of generative models.


\bibliographystyle{elsarticle-num} 
\bibliography{main}





\end{document}